\pdfoutput=1
\documentclass[11pt]{article}

\usepackage{acl}
\usepackage{amsmath, amssymb}
\usepackage[T1]{fontenc}

\usepackage{times}
\usepackage{latexsym}
\usepackage[utf8]{inputenc}
\usepackage{microtype}
\usepackage{graphicx}
\usepackage{multirow}
\usepackage[textsize=scriptsize,disable=True]{todonotes}
\title{Can Unsupervised Knowledge Transfer from Social Discussions\\ Help Argument Mining?}

\author{Subhabrata Dutta\thanks{*Equal contribution} \\
Jadavpur University\\
\texttt{subha0009@gmail.com}
\And
Jeevesh Juneja\footnotemark[1] \\
Delhi Technological University\\
\texttt{creativityinczenyoga@gmail.com}
\AND
Dipankar Das\\
Jadavpur University\\
\texttt{dipankar.dipnil@gmail.com}
\And
Tanmoy Chakraborty\\
IIIT-Delhi\\
\texttt{tanmoy@iiitd.ac.in}}

\date{}

\begin{document}
\maketitle
\begin{abstract}
Identifying argument components from unstructured texts and predicting the relationships expressed among them are two primary steps of argument mining. The intrinsic complexity of these tasks demands powerful learning models. While pretrained Transformer-based Language Models (LM) have been shown to provide state-of-the-art results over different NLP tasks, the scarcity of manually annotated data and the highly domain-dependent nature of argumentation restrict the capabilities of such models. In this work, we propose a novel transfer learning strategy to overcome these challenges. We utilize argumentation-rich social discussions from the {\em ChangeMyView} subreddit as a source of unsupervised, argumentative discourse-aware knowledge by finetuning pretrained LMs on a selectively masked language modeling task. Furthermore, we introduce a novel prompt-based strategy for inter-component relation prediction that compliments our proposed finetuning method while leveraging on the discourse context. Exhaustive experiments show the generalization capability of our method on these two tasks over within-domain as well as out-of-domain datasets, outperforming several existing and employed strong baselines.\footnote{We release all code, models and data used at \url{https://github.com/Jeevesh8/arg_mining}}

\end{abstract}

\section{Introduction}
\label{sec:intro}
Computational argument mining from texts is the fine-grained process of understanding opinion dynamics. In the most fundamental sense, argument understanding requires the identification of the opinions posed and justifications provided to support or falsify them. Generally, automated argument mining is a multi-stage pipeline identified with three general steps \citep{AM-ML-survey, Stab-Gurevych-2017-Parsing-Persuasive} -- separating argumentative spans from non-argumentative ones, classifying argument components, and inducing a structure among them (support, attack, etc.). 
While different argumentation models define different taxonomies for argument components, popular approaches broadly categorize them as  `claims' and 
`premises' \citep{Stab-Gurevych-2017-Parsing-Persuasive, egawa-etal-2019-annotating, bert-gru-crf}. As these components are not necessarily aligned to sentence-level segments and can be reflected within clausal levels, the task of argument component identification requires a {\em token-level boundary detection} of components and {\em component type classification}. 

{\bf Context of argumentation in online discussions.} 
\begin{figure}[!t]
    \centering    \includegraphics[width=\columnwidth]{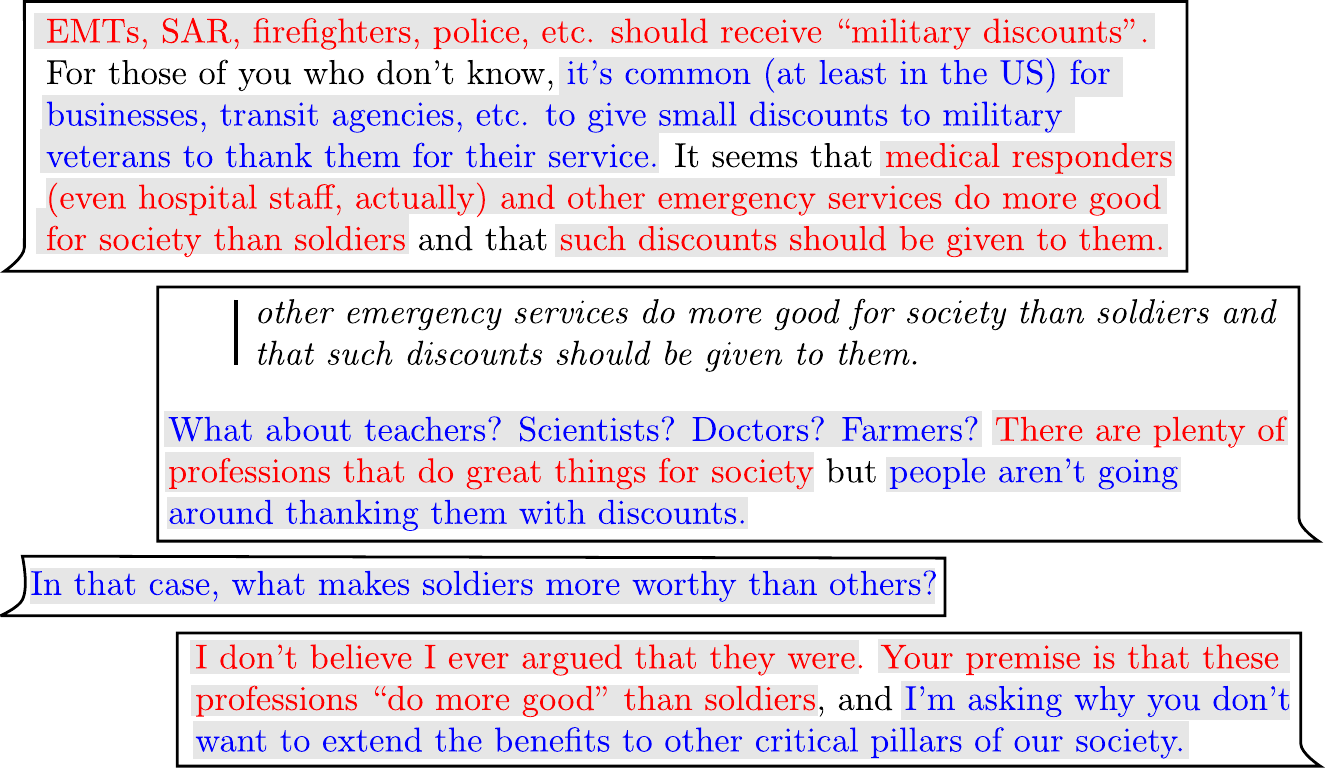}
    \caption{\footnotesize Token-level claim (red) and premise (blue) annotation of a discussion thread formed by consecutive posts from two users. Second post quotes a span from the first (shown in italics). Highlighted regions signify component boundaries (to demarcate consecutive components of the same kind as in the fourth post).}
    \label{fig:intro-example}
    \vspace{-5mm}
\end{figure}
Online discussions originating from back-and-forth posts from users reflect a rich interaction of opinion dynamics on large scale. In Figure~\ref{fig:intro-example}, we show a sample argument component annotation of consecutive posts from two users. The token-level granularity of components ensures that a single sentence may contain multiple components of the same (in 1st post) or different kinds (in 2nd and 4th posts). Moreover, two adjacent spans of texts, even with the same argumentative role, can be defined as two separate components (see the 4th post for example). It is trivial to say that the meaning of any post (as well as its argumentative role) is dependent on the context. To be specific, the third post can be identified as argumentative (a premise in this case) only when its predecessor post and its components are taken as the context. Similarly, a certain span of the first post is quoted in the second one signaling a concrete manifestation of dialogic continuity. One may even observe the user-specific argumentation styles: 1st user (author of the first and third posts) usually keeps claims and premises in separate sentences, while the 2nd user prefers to use multi-component, complex sentences. Existing studies on argumentation formalism recognize such continuity and define inter-post component relations \citep{ghosh-etal-2014-analyzing, dataset-original}. However, the previous approaches for automated extraction, classification and relating argumentative components work on individual posts only and define the inter-post discourse in the later stages of relation prediction.

This is trivially counter-intuitive for two major reasons: (i) if we consider two text spans from separate comments to be linked by some argumentative relation, then there exists a continuity of discourse between these spans and a model is likely to benefit if it decides the boundaries and types of these two components conditioned on that continuous information; (ii) users carry their style of argumentation (simple consecutive sentences vs. long complex ones, usage of particular markers like `{\em I think that}' etc.), and if the model is informed about these while observing the complete conversation with back-and-forth posts, it is more likely to extract correct components easily. 

{\bf Scarcity of labeled data.} Irrespective of the domain, argument annotation is a resource-intensive process. A few previous studies \citep{habernal-gurevych-2015-exploiting, al-khatib-etal-2016-cross} attempted to exploit a large amount of unlabeled data in a semi-supervised fashion. However, such methods require the components to be defined at sentence-level (and thereby adding redundant spans into the predictions) as they perform some sentence similarity matching to generate pseudo-labels. 
Pretrained language models like BERT \citep{devlin-etal-2019-bert} provide a workaround to handle the scarcity of task-specific annotated data. A parameter-intensive model is initially trained in a self-supervised manner on a large bulk of text; this pretraining enables the model to learn general language representation, which is then finetuned on task-specific labeled data. However, the amount of the latter still determines the expressive power of such models \citep{pretrain-gap}. 

{\bf Present work.} Considering these challenges, we formulate a novel transfer learning method using Transformer-based language models. We use large amount of unlabelled discussion threads from Reddit's {\em r/ChangeMyView} (CMV) community as the source of argumentative knowledge. Pretrained, Transformer-based language models are finetuned on this dataset using a Masked Language Modelling task. Instead of randomly masking tokens to predict, we select several markers in the text that are shown to signal argumentative discourse in previous works~\cite{ampersand, dm-role}. The language models are then made to predict these markers in the MLM task, thereby learning to relate different components of text according to their role in the argumentation presented. We call this novel finetuning method Selective Masked Language Modeling ({\bf sMLM}). Furthermore, to explore the role of context in argument mining, we use {\bf sMLM} to finetune a post-level language model based on BERT~\cite{devlin-etal-2019-bert} and RoBERTa~\cite{liu2019roberta} and a thread-level language model based on Longformer~\cite{longformer}. We present efficient incorporation of several Reddit-specific structural cues into the Longformer architecture. These finetuned language models are then used for two fundamental components of argument mining: token-level argument component identification (\textbf{ACI}) and inter-component relation type prediction (\textbf{RTP}). To further utilize the {\bf sMLM}-based training of the language models, we propose a novel prompt-based approach to predict relations among argument components. We perform exhaustive experiments to explore the efficacy of our proposed methods for argument mining in both {\em in-domain} and {\em out-of-domain} benchmark datasets: manually annotated Reddit discussions and scientific papers. Our experiments show clear improvements achieved by our  methods ($0.59$ and $0.69$ F1 for {ACI} and {RTP}, respectively) over several state-of-the-art baselines.\footnote{The source codes and datasets have been submitted separately.}

\section{Related Work}
\label{sec:rw}

A general overview of argument mining can be found in the survey articles by \citet{LYTOS2019102055} and  \citet{lawrence-reed-2019-argument}. In the current scope, we look into three major areas of research in argument mining.

{\bf Argument component detection and classification.} Previous studies have sought to address argument boundary detection and component type prediction either as separate, successive tasks in the pipeline \citep{Stab-Gurevych-2017-Parsing-Persuasive} or jointly in a single computational pass \citep{NeuralE2E}. Studies also explored classical machine learning frameworks like SVM-HMM \citep{Habernal-Gurevych-2017-Web}, CRF \citep{Stab-Gurevych-2017-Parsing-Persuasive}, etc. with rich manual feature engineering. With the development of neural network-based algorithms, BiLSTM-CNN-CRF models emerged as a popular choice \citep{mtl-am, NeuralE2E, targer}. Very recently, large pretrained language models like BERT have also been utilized \citep{bert-gru-crf, ampersand}.

%{\bf Data scarcity in argument mining.} \citet{mtl-am} sought to generalize argument mining to handle the scarcity of annotated datasets by exploiting a multi-task learning scenario on multiple datasets with different annotation schemes. \citet{DUTTA2020102085} defined some rules to identify sentences containing claims and premises with high precision from Reddit comments and used those as seed components to cluster undetected components. \citet{shnarch-etal-2018-will} augmented weakly-labeled data with gold-standard ones using neural networks to identify topic-dependent evidence from the text. \citet{habernal-gurevych-2015-exploiting} proposed a semi-supervised architecture for claim-premise identification, which uses gold-standard data augmented with texts from debate portals. They performed argument segmentation at sentence-level instead of token-level, thereby relaxing some complexities. \citet{al-khatib-etal-2016-cross} took similar strategy to classify text spans into argumentative/non-argumentative.

{\bf Discourse markers for learning language representation.} Similar to our {\bf sMLM} finetuning strategy, \citet{nie-etal-2019-dissent} proposed an unsupervised sentence representation learning strategy where a neural model is trained to predict the appropriate discourse marker connecting two input sentences. Using a set of 15 markers, they showed that such a finetuning can help models in downstream NLI tasks. \citet{ampersand} used a distant supervision approach using a single marker {\em In my honest opinion} to finetune BERT on a large collection of {\em ChangeMyView} threads and then performed argument component classification. However, they did not deal with the component identification task and performed classification of already identified components at sentence-level. \citet{opitz-frank-2019-dissecting} suggested that while identifying the relation between two components, these models often rely more on the context and not the content of the components; discourse markers present within the context provide strong signals for the relation prediction task. 

{\bf Argument mining over Reddit.} A few recent studies explored argumentation over Reddit. \citet{dataset-original} proposed a two-tier annotation scheme of claim-premise components and their relations, defining five different semantic roles of premises, using {\em ChangeMyView} discussion data. \citet{egawa-etal-2019-annotating} also analyzed semantic roles of argument components over {\em ChangeMyView} threads; however, their primary focus remained on the dynamics of persuasion, similar to \citet{DUTTA2020102085}. 

\section{Selective MLM finetuning of Pretrained Language Models}
\label{sec:sMLM-method}

Though pretrained language models are developed to overcome the problem of small annotated data on different language processing tasks, they still require task-specific finetuning for better results~\citep{pretrain-gap}. In the specific domain of argument mining, annotated data is scarce, and attempting to finetune a massive language model with very small training data comes with the risk of overfitting. Moreover, different datasets follow different strategies for annotation. We seek to devise a novel transfer learning strategy where a given Transformer-based pretrained language model is directed to focus on argumentative discourse using large-scale, unlabelled data. We choose the {\em ChangeMyView} (CMV) community as the source of this transfer for two specific reasons: (i) it provides us with a large, readily available resource of interactions strictly focused on debates around versatile topics, and (ii) discussions in CMV contain a mixture of dialogic continuity over successive turns along with elaborate argumentation presented in a single turn. We hypothesize that such a versatile combination of discourse can make the language model more generalizable over dialogic as well as monologic argument mining tasks.

\subsection{Discourse structure of CMV}
\label{subsec:prelims}

Discussion forums like Reddit facilitate users to begin a discussion with an initial post ({\em submissions},  in the case of Reddit) and then comments under that post to instantiate a discussion. Users may post a comment in reply to the submission as well as the already posted comments. A typical discussion over Reddit forms a tree-like structure rooted at the submission. Any path from the root to a leaf comment can be perceived as an independent dialogic discourse among two or multiple users; henceforth, we will call such paths as {\em threads}. Formally, a thread $T$ is an ordered sequence $\{(u_i, P_j)|i,j\in \mathbb{N}, u_i\in U_{T}\}$, where $P_j$ is a text object (a submission when $j=1$ and a comment, otherwise), $u_i$ is the author of $P_j$, and $U_{T}$ is the set of all unique users engaged in the thread $T$. For brevity, we indicate $P_j$ as a post in general. 

The dialogic nature of discussions naturally assumes this context to be the whole thread $T$. However, if we consider any two successive posts $P_j$ and $P_{j+1}$ in $T$, they manifest the interests and styles of two separate participants along with the discourse continuity of the overall thread, which must be distinguished within the definition of the context. 
To take into account the complete dialogic context of the thread, we represent a thread as a single contiguous sequence of tokens with each post $P_j$ from user $u_i$ being preceded by a special token $[\mathbf{USER}\text{-}i]$ with $i\in \{0, \cdots, |U_T|-1\}$, to encode which post is written by which user. 

Reddit also offers users a {\em quoting} facility: users can quote a segment from the previous post (one to which they are replying) within their posts and emphasize that their opinions are specifically focused on that segment. We delimit such quoted segments with special tokens {\bf [STARTQ]} and {\bf [ENDQ]} in the quoting post to demarcate the dialogic discourse. \citet{ampersand} also used quoting as signals for following premises. Additionally, we replace URLs with the  special token {\bf [URL]} to inform the presence of external references that often act as justifications of subjective opinions.

\subsection{Selective MLM finetuning}
\label{subsec:selective-mlm}

Masked Language Modeling is a common strategy of training large language models; a certain fraction of the input tokens are masked and the model is trained to predict them, consequently learning a generalized language representation. Instead of randomly selecting tokens to mask, we select specific markers that might signal argumentative discourse. While the model is trained to predict these markers, it learns the roles and relationships of the text spans preceding and following them. Following the work by \citet{dm-role}, we select multiple markers signaling {\em Opinion}, {\em Causation}, {\em Rebuttal}, {\em Fact presentation}, {\em Assumption}, {\em Summary}, and some additional words, which serve multiple purposes depending on the context. 

\begin{figure}
    \centering
    \includegraphics[width=\columnwidth]{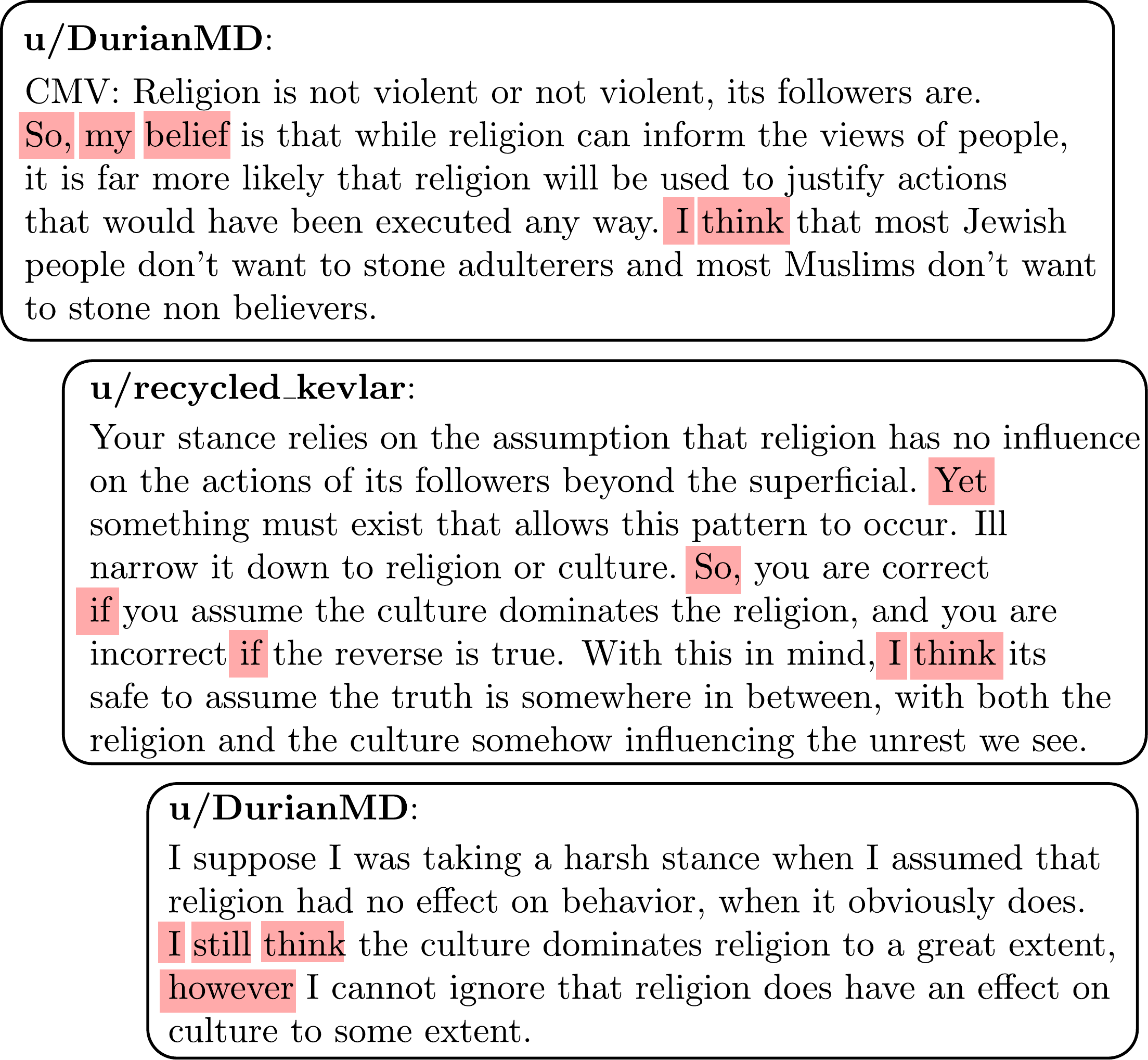}
    \caption{\footnotesize Example of selective masking in a sample CMV thread; {\bf sMLM} finetuning requires a pretrained language model to predict the masked (highlighted in red) tokens (or all the subwords constituting them) based on the context.}
    \label{fig:sMLM-example}
\end{figure}

As shown in Figure~\ref{fig:sMLM-example}, to predict the marker {\em I think} in the first post, the model needs to learn that the following text span ``{\em that most Jewish people $\cdots$}'' expresses the user's opinion on the topic. Similarly, in the second post, for the input segment ``$\langle span_0\rangle$ {\em So} $\langle span_1\rangle$ {\em if} $\langle span_2\rangle$'', to correctly predict the masked markers as {\em So} and {\em if}, a language model needs to learn the fact that the truth value of the statement expressed in $\langle span_1\rangle$ is conditioned upon $\langle span_2\rangle$, and this dependence is inferred from $\langle span_0\rangle$.

%Formally, given a thread $T$ represented as a sequence of tokens (after the usual noise removal steps like omitting special characters, handling unicode characters, removing bot responses, etc. and the special token insertions discussed in Section~\ref{subsec:prelims}) $\{w^k_T|w^k_T\in \mathbf{V}\}_{k=1}^{N}$ of length $N$ with fixed token vocabulary $\mathbf{V}$, and a set of markers $\mathbf{M}{:=}\{m_i|m_i{=}\{w^k_i\}_{k=1}^{|m_i|}, w^k_i\in\mathbf{V}\}$, the input for the selective MLM step can be defined as $X=\{x^k|x^k{=}w_m\text{ if }w^k_T{\in}m_i{\in}\mathbf{M}, w^k_T \text{ otherwise}\}_{k=1}^{N}$, where $w_m$ is a special token $\langle mask\rangle$. 

%A pretrained language model computes $h^k\in \mathbb{R}^{d_\text{LM}}$ for each $x^k\in X$, $d_\text{LM}$ being the model dimension:
%\begin{equation}
%\label{eq:token-level-rep}
%\small
%    h^k = \text{LM}(x^k|X; \theta_{\text{LM}})
%\end{equation}
%from which a {\em language model head} computes softmax probability distributions over %$\mathbf{V}$ as follows:
%\begin{equation}
%\small
%    p(\Tilde{x}^k=v\in \mathbf{V}) = \operatorname{softmax}(W_{\text{emb}}^\top h^k + B)
%\end{equation}
%where $W_{\text{emb}}\in \theta_{\text{LM}}$ is the embedding matrix of the model, and $B$ is a bias matrix. The usage of transposed embedding matrix to project the hidden representation back to the vocabulary space is inspired by \citet{tied-embedding}.

%We optimize $\theta_{\text{LM}}$ by minimizing the cross-entropy loss between the predicted and actual tokens in the masked position.

\paragraph{Effect of context sizes.} CMV threads provide a natural segmentation of the discourse context into comment/post-level vs. thread-level. We seek to explore the effect of the context size at different modules of argument mining (i.e., argument component detection and relation type prediction). For this, we use our proposed selective MLM approach to finetune a pretrained RoBERTa/BERT-base model in the comment/post-level regime, and train Longformer models in the thread-level regime. Longformer uses sparse, global attention (i.e., some tokens attend to all the tokens in the input sequence) to capture the long-range dependencies. We use the special tokens indicating the users (c.f. Section~\ref{subsec:prelims}) as the globally attending tokens for Longformer. 

\subsection{Argument component identification}
\label{subsec:arg-comp-identify}

After finetuning the language model on the selective MLM task, we proceed to our first task of identifying argument components in threads. Since the detection is done in token-level, we use the standard {\bf BIO} tagging scheme: for a component class $\langle type \rangle$, the beginning and the continuation of  that component are marked as $\mathbf{B}$-$\langle type \rangle$ and $\mathbf{I}$-$\langle type \rangle$, respectively, while any non-component token is labeled as $\mathbf{O}$. Therefore, if one uses the usual claim-premise model of argumentation, the label set becomes $\{\mathbf{B}\text{-claim}, \mathbf{I}\text{-claim}, \mathbf{B}\text{-premise}, \mathbf{I}\text{-premise}, \mathbf{O}\}$. 

\subsection{Inter-component relation prediction}
\label{subsec:method:relation-prediction}

% In this task, our aim is to predict the relation type between various argumentative components identified during segmentation. 
% We add an extra "continue" class of relations to denote relation between two dis-contiguous spans of same argumentative component annotated in the data. We do a 5 way classification of relations into:
% \textbf{support}, \textbf{agreement}, \textbf{direct attack}, \textbf{undercutter attack}, \textbf{partial} classes. 

\begin{figure}
    \centering
    \includegraphics[width=\columnwidth]{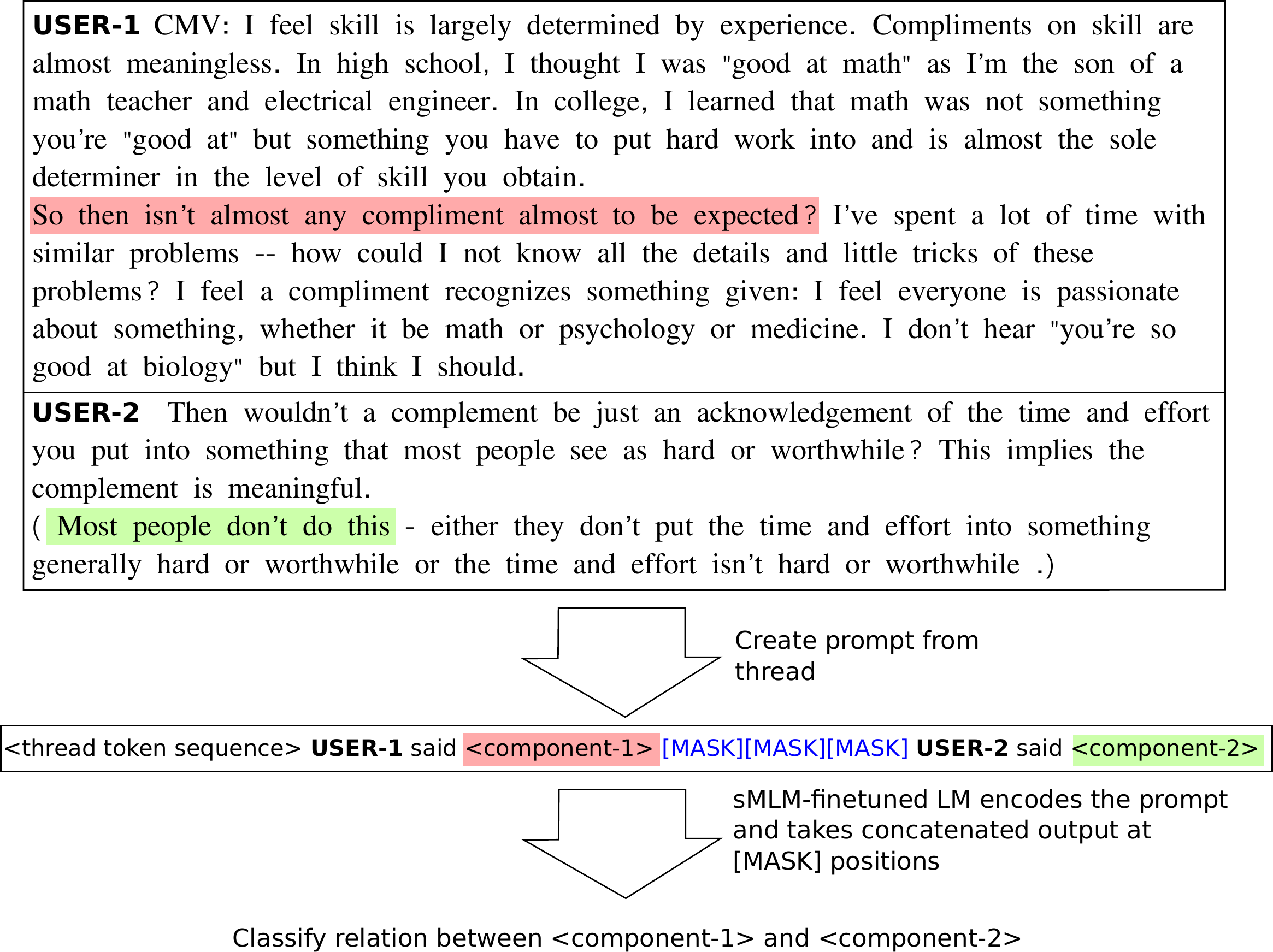}
    \caption{\footnotesize Example outline of prompt-based relation prediction where we seek to classify the relation between the claims posed by {\bf USER-1} and {\bf USER-2}, highlighted in red and green, respectively; the thread is converted to the prompt input by appending the prompt template. The language model the converts this prompt token sequence into fixed dimensional vectors from which the vector corresponding to the position of the masking token is used for relation classification.}
    \label{fig:method-prompt}
\end{figure}

While identifying the relation between two given related argument components, it is important to understand the role of those text segments within the context of the discourse. Furthermore, we seek to utilize the knowledge acquired by a language model in the {\bf sMLM} finetuning step as well. Keeping these two factors in mind, we propose a novel, prompt-based identification of argument components. This approach is inspired by recent popularity of prompt-based fine-tuning methods in the community~\cite{liu2021pretrain}. At its core, these methods involve directly prompting the model for the required knowledge, rather than fine-tuning {\bf [CLS]} or mean-pooled embeddings. For example, to directly use a model to summarise a text, we can append "{\bf TL;DR:}" to the text~\cite{Radford2019LanguageMA}, and let the model generate tokens following it; we expect the next few tokens to constitute a summary of all the previous text. 

Since the underlying Transformer LMs have been trained using some Cloze task\todo{what is cloze} (i.e., filling the blanks from the context) previously, it is more natural for it to predict a token given a context. However, there are two challenges: (i) one needs to design a suitable prompt, and (ii) in case of classification tasks like {\bf RTP}, it is challenging to perform {\bf Answer Mapping}, i.e., to map all the possible tokens to some particular relation class. 
%Tokens which might signal a particular type of relation in a certain domain may not fit in some other domain. 
To tackle these challenges, we design our proposed relation prediction method in the following manner (see Figure~\ref{fig:method-prompt})

For each pair of related components, say, component-1 and component-2, said by user-i and user-j, respectively, where component-2 refers to component-1, we append to the thread, a prompt with the template: "[USER-i] said <component1> [MASK] [MASK] [MASK] [USER-j] said <component2>" (we used three mask tokens since that is the upper bound of the marker size used for {\bf sMLM}). We expect that the words predicted at the masked position such as ``because'', ``in spite of what'' etc. would be indicative of the relation of the two components. For the example thread shown in Figure~\ref{fig:method-prompt}, in a zero-shot prediction, {\bf sMLM}-finetuned Longformer predicts "I", "disagree", "I" at the three masked positions\todo{check this line}. This ``disagree" clearly corresponds to the \textbf{undercutter} relation between the two components. In fact, the base Longformer without {\bf sMLM} finetuning predicts a space, a full stop and another space at the three masked positions. This additionally proves the efficacy of the {\bf sMLM} finetuning.

Instead of engineering a token-to-relation type mapping, the predicted token embeddings at the masked positions are concatenated and fed into a linear layer to predict probabilities over the set of relation types. This way, we allow the model to learn and map from the token space to the relation type space. \todo{In general, this para needs polishing, very differently/awfully written from the other para}

\section{Experiment Setup}
\label{sec:experiments}
 
\subsection{Dataset}
\label{subsec:data}

For the {\bf sMLM} finetuning, we use the subset of {\em Winning Args (ChangeMyView)} (CMV) dataset \citep{MLMfinetune-data} provided in ConvoKit \citep{chang2020convokit}. We use 99\% of this data for training, and reserve 1\% for checking accuracy on the {\bf sMLM} task. The entire data consists of $3,051$ submissions and $293,297$ comments posted in the {\em ChangeMyView} subreddit by $34,911$ unique users. We extract the threads from these posts following the reply structure and end up with $120,031$ threads in total.

To train and evaluate all the models for \textbf{ACI} and \textbf{RTP}, we use the manually annotated Reddit discussion threads provided by \citet{dataset-original} and further extended by \citet{ampersand} for training and evaluation. The extended version of this dataset contains $113$ CMV discussion threads manually annotated with argument components following the standard claim-premise model. 

Additionally, we use the argument annotated {\bf Dr. Inventor Corpus}~\cite{ArgDrInvCorpus} which consists of 40 scientific publications from the field of computer graphics. There are three types of argumentative components here: Background Claims (\textbf{BC}), consisting of claims from previous works in the paper, Own Claim (\textbf{OC}) consisting of the new claims made by the authors of the paper, and {\bf Data}. The Data class mainly consists of citations, references to figures, etc. This dataset has three relation types, viz., \textbf{support}, \textbf{contradicts} and \textbf{semantically same}. Additional dataset details are provided in Appendix~\ref{app:data}.

\subsection{Baseline methods}
\label{subsec:baselines}
For {\bf ACI}, we consider two state-of-the-art token-level argument identification models:
$\rhd$  \textbf{LSTM-MTL.} \citet{NeuralE2E} proposed an end-to-end argument mining architecture which uses a  BiLSTM-CNN-CRF sequence tagger to jointly learn component detection, classification, and relation parsing tasks. 
$\rhd$  \textbf{LSTM-MData.} \citet{mtl-am} proposed a BiLSTM-CNN-CRF based model which aims to generalize argument mining using multi-domain training data in an MTL setting. We augment our data with their original set of $6$ datasets.

For {\bf RTP}, as no prior work exists to the best of our knowledge, we consider our own baselines. First, we consider $\rhd$ \textbf{Context-less RoBERTa}, a pretrained RoBERTa model, which takes the two components with a [SEP] token between them and predicts the relation using [CLS] token's embedding. It is context-less as only two components without the surrounding context are used to predict the label. Second, we consider $\rhd$ \textbf{Contextless QR-Bert.} This uses the same fine-tuning methodology as \textbf{Contextless RoBERTa} and is initialized from the pre-trained Quote-Response relation prediction model of \citet{ampersand}. 

For {\bf RTP}, we try another traditional strategy, instead of prompting, for our models: $\rhd$ \textbf{Mean Pooling}. The mean pooling approach first finds an embedding of each of the two related components by averaging the Transformer embeddings at all token positions within a component. These embeddings are concatenated and passed into a linear layer for predicting the type of relation between the two related components. 
%As we use Longformer, all the tokens in a thread are embedded at once, and hence each component's embedding has context information encoded in it. The losses corresponding to each pair of related components in a thread, are summed and back-propagated at once. 

To further evaluate the efficacy of our {\bf sMLM} training strategy, we finetune a pretrained Longformer on the Winning Args Corpus, with the usual MLM, i.e., masking 15\% of tokens at random, instead of selective masking. We call this the domain-adapted Longformer, {\bf DA-LF}.

\subsection{Implementation details}
\label{subsec:implement}

We use the pretrained base version of Longformer ($12$ layers, $768$ model size). The size of the local attention window was set to the default $512$. The maximum sequence length was fixed at $4096$. 
%We added the special tokens that we used, to the pretrained Longformer tokenizer. For \textbf{ACI} our models use a CRF layer\footnote{We use the implementation of AllenNLP~\citep{gardner2018allennlp}}. \textbf{sMLM} training for Longformer based models was done on thread level and for BERT and RoBERTa based models on comment-level. We used mini-batch learning; approximately similar length input threads were batched together keeping the total number of tokens per batch fixed to $8,194$ for Longformer and $1024$ for BERT and RoBERTa models, and accumulated gradients over 3 batches. 

%We trained our models for a total of $10$ epochs on {\bf sMLM} task, while saving checkpoints after each epoch. We used Adam optimizer with a learning rate of $10^{-6}$. For all downstream tasks, we train our models for $30$ epochs, again, with Adam optimizer with learning rate of $2e-5$ as suggested by \citet{mosbach2021stability}. We use same batch sizes as \textbf{sMLM} training and accumulate gradients over 4 batches. We observe that for prompting {\bf RTP} on CMV Modes, not making {\bf [USER-i]} tokens global, leads to better performance, hence we report results for same.
Following the suggestions in \citet{reimers2017reporting}, we repeat our experiments on the 5 different data splits. The scores reported in the tables for various models correspond to the average value of the mean of 5 runs, over the last 5 epochs for that particular metric. We provide additional implementation details in Appendix~\ref{app:implement}.

\section{Evaluation}

We evaluate the models based on precision, recall, and F1 scores for predicting claims and premises. For a more rigorous setting, we use exact match of the whole span between gold and predicted labels, i.e., if the gold label is [\textbf{O}, \textbf{B}{-}claim, \textbf{I}{-}claim, \textbf{I}{-}claim, \textbf{I}{-}claim, \textbf{O}] then only the predictions [\textbf{O}, \textbf{B}{-}claim, \textbf{I}{-}claim, \textbf{I}{-}claim, \textbf{I}{-}claim, \textbf{O}], or [\textbf{O}, \textbf{I}{-}claim, \textbf{I}{-}claim, \textbf{I}{-}claim, \textbf{I}{-}claim, \textbf{O}] can be considered as true positives. We use the popular SeqEval~\citep{seqeval} framework. 

\subsection{Argument component identification}
\begin{table}[!t]
\scriptsize
    \centering
    \setlength{\tabcolsep}{2pt} % Default value: 6pt
    \begin{tabular}{l|ccc|ccc|c|c}
    \hline
    \multirow{2}{*}{\bf Model} & \multicolumn{3}{c}{\bf Claim} & \multicolumn{3}{c|}{\bf Premise} & \multirow{2}{*}{\bf F1} & \multirow{2}{*}{\bf Acc}\\ \cline{2-7} 
                       & P & R & F1  & P& R & F1  & & \\ \hline
    
    %%BEGIN NEW ADDITIONs/UPDATES
    % \multicolumn{8}{c}{\textbf{\ \ \ \ \ \ \ \ \ \ \ \ \ \ \ 80-20 split}}\\
    % \hline
    
    \multicolumn{1}{l|}{sMLM-LF}     & $0.49$ & $ 0.57$ & $\bf 0.53$ & $\bf 0.61$ & $\bf 0.67$ & $\bf 0.64$ & $\bf 0.59$ & $\bf 0.74$\\
    \multicolumn{1}{l|}{Base-LF}      & $\bf 0.50$ & $0.50$ & $0.50$ & $0.58$ & $0.64$ & $0.61$ & $0.56$ & $\bf 0.74$\\
    % \multicolumn{1}{l|}{sMLM-256-LF}  & $0.48$ & $0.54$ & $0.51$ & $0.58$ & $0.66$ & $0.62$ & $0.57$ & $0.72$\\
    \multicolumn{1}{l|}{sMLM-RoBERTa} & $0.49$ & $\bf 0.60$ & $\bf 0.53$ & $0.55$ & $0.57$ & $0.55$ & $0.55$ & $0.72$\\
    \multicolumn{1}{l|}{RoBERTa}     & $0.49$ & $0.55$ & $0.51$ & $0.56$ & $0.62$ & $0.59$ & $0.56$ & $0.73$\\
    % \multicolumn{1}{l|}{sMLM-RoBERTa} & $0.49$ & $0.60$ & $0.53$ & $0.55$ & $0.57$ & $0.55$ & $0.55$ & $0.72$\\
    \multicolumn{1}{l|}{BERT}        & $0.21$ & $0.25$ & $0.23$ & $0.19$ & $0.26$ & $0.22$ & $0.22$ & $0.62$\\
    \multicolumn{1}{l|}{LSTM-MData}  & $0.19$ & $0.18$ & $0.18$ & $0.26$ & $0.23$ & $0.24$  &  $0.22$ & $0.54$\\
    % \multicolumn{1}{l|}{sMLM-BERT}   & $0.22$ & $0.28$ & $0.24$ & $0.17$ & $0.23$ & $0.20$ & $0.21$ & $0.64$\\
    \multicolumn{1}{l|}{LSTM-MTL}    & $0.19$ & $0.18$ & $0.18$ & $0.24$ & $0.25$ & $0.24$ & $0.21$ & $-$\\
    \hline
    %     \multicolumn{8}{c}{\textbf{\ \ \ \ \ \ \ \ \ \ \ \ \ \ \ 50-50 split}}\\
    % \hline
    % \multicolumn{1}{l|}{sMLM-RoBERTa} & $0.49$ & $0.54$ & $0.51$ & $0.58$ & $0.66$ & $0.61$ & $0.57$ & $0.72$\\
    % \multicolumn{1}{l|}{RoBERTa}  & $0.44$ & $0.52$ & $0.48$ & $0.56$ & $0.63$ & $0.59$ & $0.54$ & $0.71$\\
    % \multicolumn{1}{l|}{Base-LF}  & $0.46$ & $0.52$ & $0.48$ & $0.54$ & $0.63$ & $0.58$ & $0.54$ & $0.70$\\
    % \multicolumn{1}{l|}{sMLM-256-LF}  & $0.44$ & $0.47$ & $0.45$ & $0.50$ & $0.62$ & $0.55$ & $0.51$ & $0.70$\\
    % \multicolumn{1}{l|}{sMLM-LF}      & $0.42$ & $0.42$ & $0.42$ & $0.49$ & $0.65$ & $0.55$ & $0.50$ & $0.68$\\
    % \multicolumn{1}{l|}{sMLM-BERT}  & $0.22$ & $0.26$ & $0.24$ & $0.17$ & $0.25$ & $0.20$ & $0.22$ & $0.63$\\
    % \multicolumn{1}{l|}{BERT}  & $0.19$ & $0.23$ & $0.21$ & $0.17$ & $0.25$ & $0.20$ & $0.21$ & $0.62$\\
    % \multicolumn{1}{l|}{LSTM-MData}  & $0.19$ & $0.17$ & $0.17$ & $0.23$ & $0.22$ & $0.22$ & $0.20$ & $0.53$\\
    % \multicolumn{1}{l|}{LSTM-MTL}  & $0.15$ & $0.16$ & $0.16$ & $0.18$ & $0.21$ & $0.20$ & $0.18$ & $0.53$\\
    % \hline
    \end{tabular}
    \caption{\footnotesize Performance of different models on \textbf{ACI}-task on CMV Modes dataset (P: Precision, R: Recall, F1: F1 score). The \textbf{F1} and \textbf{Acc.} in the rightmost columns denote the micro-averaged F1 score over claims and premises and the token level accuracy of predicting argumentative tags, respectively.}
    \label{tab:cmv_modes_res_comp_pred_comment_wise}
\end{table}

\begin{table}[!t]
\scriptsize
    \centering
    \setlength{\tabcolsep}{2pt}
    \begin{tabular}{l|ccc|ccc|ccc}
\hline
\multicolumn{1}{c|}{\multirow{2}{*}{\bf Model}} & \multicolumn{3}{c|}{\bf BC} & \multicolumn{3}{c|}{\bf OC} & \multicolumn{3}{c}{\bf Data} \\ \cline{2-10} 
\multicolumn{1}{c|}{}                  & P      & R     & F1     & P      & R     & F1     & P      & R      & F1     \\ \hline
                       sMLM-LF         &    $0.45$    & $\bf 0.52$  &   $0.48$     & $\bf 0.39$ &  $0.45$    &  $0.42$ &    $\bf 0.50$    & $\bf 0.48$   &  $\bf 0.48$  \\
                       Base-LF         &    $\bf 0.49$    &   $0.51$    &  $\bf 0.50$   &  $0.38$  & $\bf 0.50$  & $\bf 0.43$   & $0.44$    &  $0.44$ &  $0.44$  \\ \hline
\end{tabular}
    \caption{\footnotesize Results on Dr. Inventor dataset for argument component identification using {\bf sMLM}-finetuned and base Longformer models. 
    }
    \vspace{-5mm}
    \label{tab:dr_inv_res_comp}
\end{table}

% \begin{table}[!t]
%     \scriptsize
%     \centering
%     \setlength{\tabcolsep}{3pt}
%     \begin{tabular}{c|c|c|c}
%         \hline
%         {\bf Class}                 &{\bf Metrics}   & {\bf Base-LF} & {\bf sMLM-LF} \\ \hline
%         \multirow{3}{*}{\bf BC}         &P           &   {\bf 0.49}        &        0.45    \\ 
%                                         &R           &   0.51        &        {\bf 0.52}    \\
%                                         &F1          &   {\bf 0.50}        &        0.48    \\ \hline
        
%         \multirow{3}{*}{\bf OC}         &P           &   0.38        &       {\bf 0.39}    \\ 
%                                         &R           &   {\bf 0.50}        &       0.45    \\
%                                         &F1          &   {\bf 0.43}        &       0.42    \\ \hline
        
%         \multirow{3}{*}{\bf D}          &P           &   0.44        &       {\bf 0.50}    \\ 
%                                         &R           &   0.44        &       {\bf 0.48}    \\
%                                         &F1          &   0.44        &       {\bf 0.48}    \\ \hline
%         \multicolumn{2}{c|}{\bf Overall F1}          &   0.45        &       0.45          \\ \hline
%     \end{tabular}
%     \caption{\footnotesize Results on Dr. Inventor dataset for argument component identification using {\bf sMLM} finetuned and base Longformer models. 
%     }
%     %\vspace{-3mm}
%     \label{tab:dr_inv_res_comp}
% \end{table}

\begin{table*}[t]
\scriptsize
    \centering
    \setlength{\tabcolsep}{2pt} % Default value: 6pt
    \begin{tabular}{l|ccc|ccc|ccc|ccc|ccc|c}
    \hline
    \multirow{2}{*}{\bf Model} & \multicolumn{3}{c}{\bf Support} & \multicolumn{3}{c}{\bf Agreement} & \multicolumn{3}{c}{\bf Direct Attack} & \multicolumn{3}{c}{\bf Undercutter} & \multicolumn{3}{c|}{\bf Partial} & \multirow{2}{*}{\begin{tabular}[c]{@{}c@{}}{\bf Overall}\\ {\bf F1}\end{tabular}}\\ \cline{2-16} 
                       & P & R & F1  & P& R & F1 & P & R & F1 & P & R & F1 & P & R & F1\\ \hline
    %\textbf{\ \ 80-20 split}\\
    \multicolumn{17}{c}{\textbf{\ \ \ \ \ \ \ \ \ \ \ \ \ \ \ 80-20 split}}\\
    \hline
    \multicolumn{1}{l|}{sMLM-LF-prompt}  & $\mathbf{0.88}$ & $\mathbf{0.93}$ & $\mathbf{0.91}$ & $\mathbf{0.51}$ & $\mathbf{0.46}$ & $\mathbf{0.48}$ & $\mathbf{0.32}$ & $\mathbf{0.35}$ & $\mathbf{0.33}$ & $\mathbf{0.43}$ & $\mathbf{0.51}$ & $\mathbf{0.46}$ & $0.28$ & $0.12$ & $0.16$ & $\mathbf{0.67}$\\
    
    \multicolumn{1}{l|}{DA-LF-prompt} & $0.78$ & $0.84$ & $0.81$ & $0.44$ & $0.45$ & $0.43$ & $0.22$ & $0.19$ & $0.19$ & $0.30$ & $0.32$ & $0.30$ & $0.27$ & $0.11$ & $0.15$ & $0.61$ \\ 
    
    \multicolumn{1}{l|}{sMLM-LF-mp}  & $0.73$ & $0.87$ & $0.79$ & $0.49$ & $0.36$ & $0.38$ & $0.32$ & $0.24$ & $0.26$ & $0.32$ & $0.33$ & $0.41$ & $0.35$ & $\mathbf{0.21}$ & $\bf 0.25$ & $0.59$\\
    
    \multicolumn{1}{l|}{Base-LF-prompt}  & $0.79$ & $0.88$ & $0.84$ & $0.48$ & $0.44$ & $0.46$ & $0.30$ & $0.23$ & $0.25$ & $0.31$ & $0.39$ & $0.34$ & $\bf 0.37$ & $0.12$ & $0.17$ & $0.62$\\
    
    \multicolumn{1}{l|}{Base-LF-mp}  & $0.71$ & $0.87$ & $0.78$ & $0.47$ & $0.33$ & $0.37$ & $0.24$ & $0.17$ & $0.18$ & $0.27$ & $0.26$ & $0.26$ & $0.35$ & $0.20$ & $0.24$ & $0.56$\\
    
    \multicolumn{1}{l|}{RoBERTa}  & $0.78$ & $0.83$ & $0.80$ & $0.46$ & $0.34$ & $0.37$ & $0.29$ & $0.29$ & $0.28$ & $0.15$ & $0.24$ & $0.18$ & $0.36$ & $0.15$ & $0.20$ & $0.60$\\
    \multicolumn{1}{l|}{QR-Bert}  & $0.76$ & $0.85$ & $0.80$ & $0.46$ & $0.27$ & $0.34$ & $0.21$ & $0.13$ & $0.16$ & $0.19$ & $0.25$ & $0.20$ & $0.32$ & $0.16$ & $0.20$ & $0.59$\\
    \hline
    %\textbf{\ \ 50-50 split} \\
    
    \multicolumn{17}{c}{\textbf{\ \ \ \ \ \ \ \ \ \ \ \ \ \ \ 50-50 split}}\\
    \hline
    \multicolumn{1}{l|}{sMLM-LF-prompt}  & $\bf 0.87$ & $\bf 0.92$ & $\bf 0.89$ & $\mathbf{0.53}$ & $\mathbf{0.47}$ & $\mathbf{0.49}$ & $0.30$ & $\bf 0.28$ & $\bf 0.28$ & $\mathbf{0.45}$ & $\mathbf{0.58}$ & $\mathbf{0.50}$ & $0.35$ & $0.09$ & $0.14$ & $\mathbf{0.69}$\\
    
    \multicolumn{1}{l|}{DA-LF-prompt} & $0.85$ & $0.89$ & $0.87$ & $0.47$ & $\mathbf{0.47}$ & $0.44$ & $\mathbf{0.32}$ & $0.20$ & $0.24$ & $0.39$ & $0.55$ & $0.44$ & $0.32$ & $0.13$ & $0.16$ & $0.66$ \\ 
    
    \multicolumn{1}{l|}{sMLM-LF-mp}  & $0.70$ & $0.90$ & $0.79$ & $0.426$ & $0.22$ & $0.28$ & $0.28$ & $0.20$ & $0.22$ & $0.32$ & $0.26$ & $0.28$ & $0.38$ & $\bf 0.18$ & $0.24$ & $0.56$\\
    
    \multicolumn{1}{l|}{Base-LF-prompt}  & $0.78$ & $0.87$ & $0.82$ & $0.49$ & $0.44$ & $0.46$ & $0.30$ & $0.19$ & $0.22$ & $0.32$ & $0.40$ & $0.35$ & $0.32$ & $0.13$ & $0.18$ & $0.62$\\
    
    \multicolumn{1}{l|}{Base-LF-mp}  & $0.73$ & $0.86$ & $0.79$ & $0.36$ & $0.21$ & $0.26$ & $0.25$ & $0.18$ & $0.21$ & $0.23$ & $0.28$ & $0.25$ & $\mathbf{0.43}$ & $\bf 0.18$ & $\mathbf{0.25}$ & $0.56$\\
    
    \multicolumn{1}{l|}{RoBERTa}  & $0.72$ & $0.83$ & $0.77$ & $0.47$ & $0.25$ & $0.31$ & $0.22$ & $0.21$ & $0.21$ & $0.13$ & $0.16$ & $0.14$ & $0.17$ & $0.08$ & $0.10$ & $0.55$\\
    \multicolumn{1}{l|}{QR-Bert}  & $0.72$ & $0.84$ & $0.77$ & $0.47$ & $0.28$ & $0.34$ & $0.19$ & $0.13$ & $0.14$ & $0.13$ & $0.18$ & $0.15$ & $0.22$ & $0.07$ & $0.09$ & $0.54$\\
    \hline
\end{tabular}
    \caption{\footnotesize Relation type wise Precision (P), Recall (R) and F1 score on the  CMV Modes dataset for various models. The highest scores in every column are in \textbf{bold}. The suffix "mp" and "prompt" indicate that the model was trained using \textbf{Mean Pooling} and \textbf{Prompting} strategies, respectively. The \textbf{F1} in last column is the Micro/weighted-F1 over all the prediction classes.}
    \vspace{-3mm}
    \label{tab:rel_pred_res_cmv_modes}
\end{table*}

Table~\ref{tab:cmv_modes_res_comp_pred_comment_wise} shows the results for argument component identification on the CMV Modes dataset. We compare models based on their micro-averaged F1 over the two component types (claims, premises), and token level accuracy. Firstly, we observe huge difference in token-level accuracy scores as we move from the existing best performing LSTM based methods with accuracy of 0.54 to BERT, having an accuracy of 0.62. Such a difference is expected since pretrained language models like BERT provide a head-start in case of small datasets like CMV Modes. Though the token-level accuracy increases, the micro-averaged F1 for exact component match does not increase much till we start using RoBERTa. Since pretrained Longformer was trained originally from the RoBERTa checkpoint~\citep{longformer}, we can conclude that RoBERTa provides significant performance gain compared to BERT, owing to its larger training data and protocol.\todo{Wouldn't it be more apt to conclude gains of RoBERTa transfer to Longformer?}
\begin{table}[!t]
\scriptsize
    \centering
    \begin{tabular}{c|ccc|ccc}
\hline
\multirow{2}{*}{\begin{tabular}[c]{@{}c@{}}Relation\\ types\end{tabular}}                             & \multicolumn{3}{c|}{Base-LF-prompt} & \multicolumn{3}{c}{sMLM-LF-prompt} \\ \cline{2-7} 
                    & P    & R   & F1   & P  & R  & F1                      \\ \hline
Support             & {\bf 0.91}  & 0.90  & {\bf 0.91}  & 0.89  & {\bf 0.92} & {\bf 0.91}   \\
Contradict          & 0.60        & {\bf 0.60}  & {\bf 0.60} & {\bf 0.65}  & 0.55   & {\bf 0.60} \\
\begin{tabular}[c]{@{}c@{}}
Semantically\\ same
\end{tabular}    & 0.74  &\multicolumn{1}{c}{\bf 0.77}  &\multicolumn{1}{c|}{0.75}  & {\bf 0.77} &\multicolumn{1}{c}{0.75} & \multicolumn{1}{c}{\bf 0.77}  \\ \hline
\end{tabular}
    \caption{\footnotesize Relation Type wise Precision (P), Recall (R) and F1 score on Dr. Inventor Corpus for prompt-based relation prediction using {\bf sMLM} and base Longformer models.}
    \label{tab:rel_pred_res_dr_inv}
    \vspace{-5mm}
\end{table}
Longformer trained with our proposed {\bf sMLM} finetuning clearly outperforms the rest of the models in terms of overall F1 score for component identification. However\todo{Probably change to moreover?}, the effects of selective MLM is more prominant in case of thread-level context (i.e, Longformer) compared to comment-level context (i.e, RoBERTa). 

We can observe that context plays different roles for different component types: while {\bf sMLM}-finetuned Longformer and RoBERTa perform comparably for claim detection, in case of premises, the access to the complete context helps the Longformer to perform better. We can observe a similar trend in {\bf ACI}-task on Dr. Inventor dataset (see Table~\ref{tab:dr_inv_res_comp}). While Base Longformer performs comparable to its {\bf sMLM} counterpart to detect Background and Own Claims, {\bf sMLM} provides a 4 point improvement in F1 score for the Data class which plays a similar role of premises towards the claims. Intuitively, textual segments expressing claims contain independent signals of opinion that is less dependent on the context; pretrained language models might be able to decipher their roles without additional information either from the thread-level context (in case of CMV Modes, specifically) or enhanced relation-awareness induced by the {\bf sMLM} finetuning. However, identifying segments that serve the role of premises to a claim intrinsically depends on the claims as well as the discourse expressed in a larger context.

% \citet{NeuralE2E} observe that argument extraction and classification on Persuasive Essays \citep{persuasive_essays_data} dataset work better on paragraph level rather than essay level. Intuitively, the task becomes more difficult with bigger context. A claim in one comment, may act as a premise for another claim in another comment in the same thread, making it difficult to say whether a component is a claim or premise. Although we do observe drops in F1 for \textbf{Base-LF} and \textbf{sMLM-256-LF} on 80-20 split, when moving from comment level to thread level; these drops are not as strong and consistent as those in \citet{NeuralE2E}.

% Next we observe that our \textbf{sMLM} trained models sometimes lead to slight improvements and sometimes to slight degradations in performance on the CMV Modes dataset. We observe that there are no downsides to \textbf{sMLM} training for Longformer, under domain-shift, in Table \ref{tab:dr_inv_res_comp}; same micro F1 score is maintained for the 80-20 split.

\subsection{Relation type prediction}

In Table \ref{tab:rel_pred_res_cmv_modes}, we present the results for relation type identification on the CMV Modes\todo{make it consistent} dataset. We again compare models based on their micro-averaged F1 over all relation types. Firstly, we consider the traditional mean pooling approach. Within this approach, we observe a 3 point improvement\todo{what is this? Percentage points.. do you know something more apt?} for the {\bf sMLM} pre-trained Longformer on the 80-20 split, while maintaining same performance on the 50-50 split. %This indicates the superiority of the {\bf sMLM} embeddings for this task. 
\begin{figure}[!t]
    \centering
    \includegraphics[width=\columnwidth]{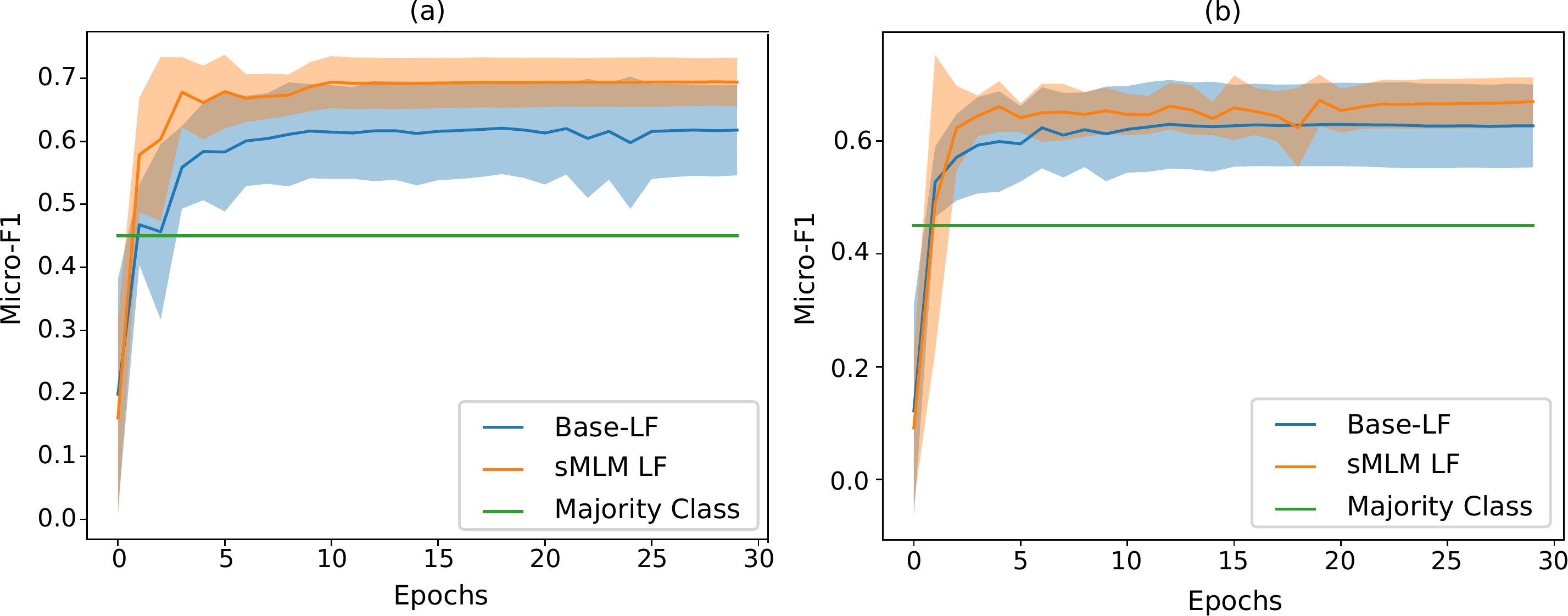}
    \caption{\footnotesize Micro-F1 scores for predicting relation types among argument components by Base and sMLM-finetuned Longformer models over the course of training using (a) 50-50 split and (b) 80-20 split. We use 5 different runs on random splits for each model to report the mean (solid lines) and variance.}
    \label{fig:RTP-comparison-plot}
    \vspace{-5mm}
\end{figure}
\todo{Fig 4: legend should be large, x/y axes large  Done}
Furthermore, the prompt based methods consistently outperform the mean pooling one, irrespective of whether we use base Longformer or {\bf sMLM} pretrained one. 

Within the prompting approach, we also observe increased and consistent improvement in performance due to {\bf sMLM} pretraining on both 80-20 and 50-50 splits. The gap in micro-F1 scores between {\bf sMLM} and base Longformer for 80-20 split increases from 3 points in mean pooling to 5 points in prompting (0 to 7 points improvements for 50-50 split). As we can observe in Figure~\ref{fig:RTP-comparison-plot}, {\bf sMLM}-finetuned Longformer admits a very narrow margin of variation on random splits, compared to the base Longformer. Furthermore, {\bf sMLM} finetuning consistently outperforms domain-adapted finetuning ({\bf DA-LF}), indicating the unique knowledge transfer achieved by the former.
%Surprisingly, the performance for 80-20 split actually turns out to be lower than that on the 50-50 split. 

We hypothesise that this approach works better as this regime models our final {\bf RTP} task, as a task that is \textit{more natural} (in a sense similar to the $(\tau, B)-$natural tasks of \citet{saunshi2021mathematical}) for a Longformer model pre-trained with {\bf sMLM}. Intuitively, the model learns to predict discourse markers at masked positions during {\bf sMLM} pre-training and during fine-tuning on downstream tasks too, the model will naturally try to predict discourse markers at the masked positions. The discourse markers occurring at the masked positions are directly related to the relation between the two components. For instance, when there is a  ``but'' between two components, we know that the two components present opposing views more or less. 
%A disadvantage of this approach is that the training time required, scales as $\mathcal{O}($no. of relations$)$ compared to $\mathcal{O}($no. of threads$)$ for the average pooling method. We present the results under domain-shift to Dr. Inventor Corpus, in Table \ref{tab:rel_pred_res_dr_inv}. 
Here again, we observe that {\bf sMLM} does not hurt the base performance under domain shift (Table~\ref{tab:rel_pred_res_dr_inv}).
\begin{figure}[!t]
    \centering
    \includegraphics[width=0.8\columnwidth]{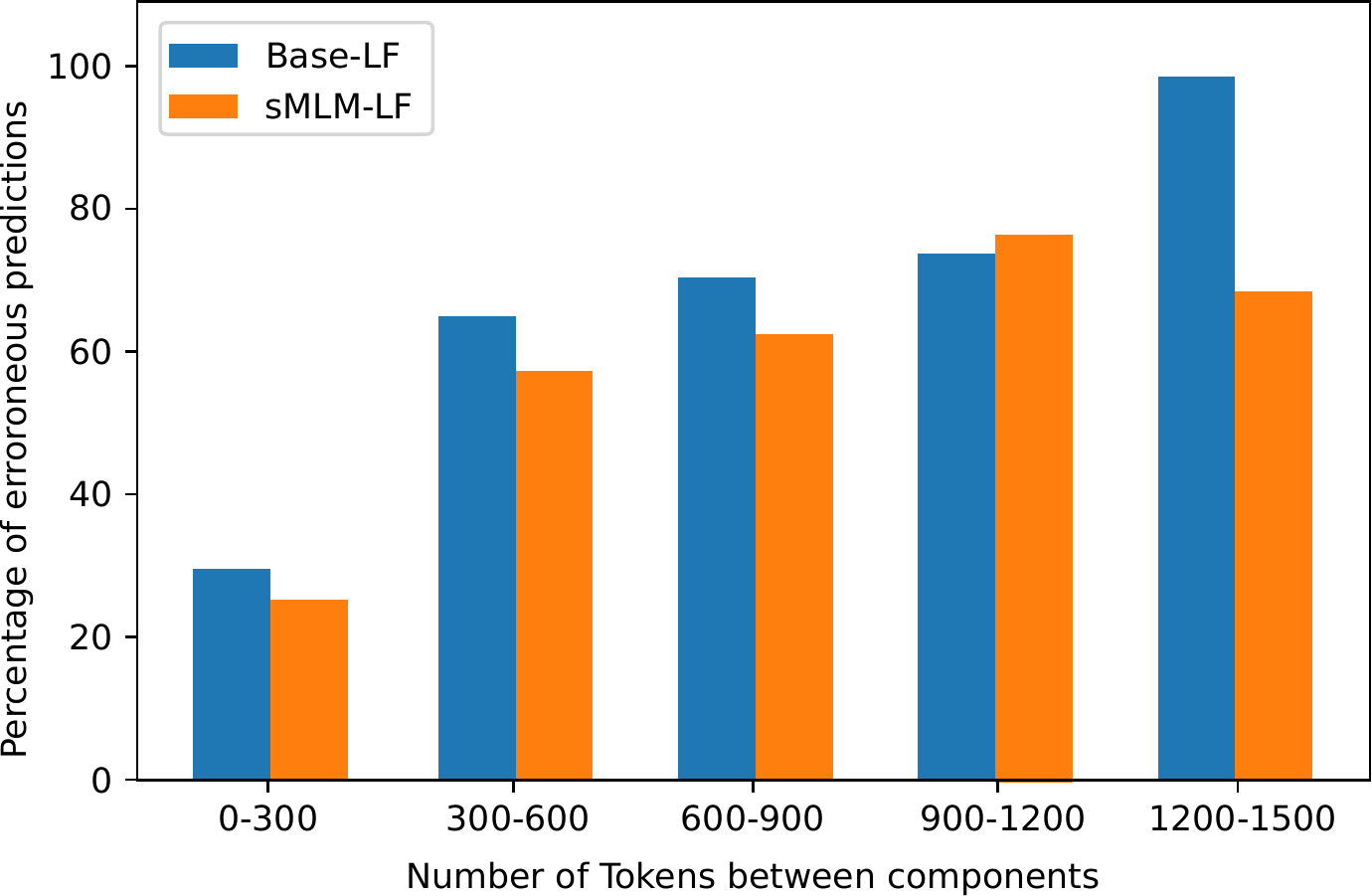}
    \caption{\footnotesize Percentage of erroneous classifications for \textbf{RTP} for Base-LF-prompt and LF-sMLM-prompt on component-pairs at different distances.}
    \label{fig:lf_rtp_error_with_dist}
    \vspace{-5mm}
\end{figure}

We observe that the RoBERTa model performs worse than Base-LF-prompt, which incorporates the entire context of the thread. Also the effect worsens with reduced training set size, and RoBERTa model performs worse by 7 points in terms of micro-F1 for the 50-50 split. Furthermore, we observe that the mean pooling strategy, even though it uses context, performs worse (by 4 points on 80-20 split) than the context-less RoBERTa. Though, our {\bf sMLM} pretrained model, manages to perform at par with the context-less RoBERTa with the mean pooling strategy. {\em This means, that the using the right fine-tuning method is essential. Extra context can be utilised fully in longformer, only when pre-training and fine-tuning tasks are nicely aligned.}

\subsection{Dependence on the presence of markers}
\label{subsec:marker-dependence}
Following the analyses presented by \citet{opitz-frank-2019-dissecting}, we investigate whether the presence/absence of the markers used in the {\bf sMLM} step within the vicinity of the components play any role in the {\bf ACI} or {\bf RTP} performances. Since the relation type among component-pairs that reside distant from each other are less likely to be inferred by the presence of markers in the context, we analyse the percentage of wrong predictions as we vary the distance between two related components, in Figure~\ref{fig:lf_rtp_error_with_dist}. While error rate does vary proportionally to the distance, we observe that {\bf sMLM-LF} consistently yields lower percentage of wrong predictions as we vary the distance between the related components compared to base Longformer. This clearly indicates the superior capability induced by the {\bf sMLM} finetuning to decipher the relationship among components not linked by direct context (i.e., not within a sentence or a single comment). 

\begin{table}[!t]
\scriptsize
    \centering
    \setlength{\tabcolsep}{2pt} % Default value: 6pt
    \begin{tabular}{l|ccc|ccc|c}
    \hline
    \multirow{2}{*}{\bf Model} & \multicolumn{3}{c}{\bf Claim} & \multicolumn{3}{c|}{\bf Premise} & \multirow{2}{*}{\bf F1}\\ \cline{2-7} 
                       & P & R & F1  & P& R & F1  \\ \hline
    
    %%BEGIN NEW ADDITIONs/UPDATES
%    \multicolumn{8}{c}{\textbf{\ \ \ \ \ \ \ \ \ \ \ \ \ \ \ 80-20 split}}\\
    \hline
    \multicolumn{1}{l|}{base-LF-near}  & $0.39$ & $0.59$ & $0.47$ & $0.63$ & $0.52$ & $0.57$ & $0.52$\\
    \multicolumn{1}{l|}{base-LF-far}  & $0.42$ & $0.57$ & $0.48$ & $0.63$ & $0.55$ & $0.59$ & $0.54$\\
    \multicolumn{1}{l|}{sMLM-LF-near}  & $0.36$ & $0.48$ & $0.40$ & $0.68$ & $0.65$ & $0.66$ & $0.57$\\
    \multicolumn{1}{l|}{sMLM-LF-far}  & $0.46$ & $0.57$ & $0.51$ & $0.63$ & $0.63$ & $0.63$ & $0.58$\\
    \hline
    % \multicolumn{8}{c}{\textbf{\ \ \ \ \ \ \ \ \ \ \ \ \ \ \ 50-50 split}}\\
    % \hline
    % \multicolumn{1}{l|}{base-LF-near}  & $0.38$ & $0.59$ & $0.46$ & $0.57$ & $0.57$ & $0.57$ & $0.52$\\
    % \multicolumn{1}{l|}{base-LF-far}  & $0.43$ & $0.59$ & $0.50$ & $0.61$ & $0.64$ & $0.63$ & $0.57$\\
    % \multicolumn{1}{l|}{sMLM-LF-near}  & $0.35$ & $0.59$ & $0.44$ & $0.55$ & $0.55$ & $0.55$ & $0.50$\\
    % \multicolumn{1}{l|}{sMLM-LF-far}  & $0.43$ & $0.60$ & $0.50$ & $0.60$ & $0.61$ & $0.60$ & $0.55$\\
    % \hline
    \end{tabular}
\caption{\footnotesize Performance of base Longformer and {\bf sMLM} Longformer for predicting segments having some markers in "\textbf{near}" (5 tokens on either side of its) boundaries, and the rest of segments ("\textbf{far}").}
\vspace{-5mm}
\label{tab:dm_closeness_wise_error}
\end{table}
For the {\bf ACI} task, however, we observe that the absence of markers in the vicinity of the components actually enables better identification, both in case of {\bf sMLM} finetuned and pretrained Longformer (see Table~\ref{tab:dm_closeness_wise_error}). 
\section{Conclusion}
We presented the results for two important tasks in the argument mining pipeline, viz., {\bf ACI} and {\bf RTP}. The experiments clearly elucidated the importance of alignment between the downstream and pre-trainig tasks, and the effect of various ways of modelling the tasks. The importance of entire thread's context in discussion forums, as well as how to incorporate that into transformer-based models fruitfully has also been made clear. %We hope that our results serve as strong baselines for future work. 
\label{sec:conclude}

\section*{Acknowledgements}
The authors would like to thank Chris Hidey and Smaranda Muresan, for clarifications providing regarding their work. T. Chakraborty would like to acknowledge the support of Ramanujan Fellowship, CAI, IIIT-Delhi and ihub-Anubhuti-iiitd Foundation set up under the NM-ICPS scheme of the Department of Science and Technology, India.

\bibliography{ref}
\bibliographystyle{acl_natbib}
\newpage
\newpage
\appendix
% \section{Appendix}
% \label{sec:Appendix}

%%DONT EDIT ONLY COPY$$
\section{Dataset Details}
\label{app:data}

Stats for the CMV Modes dataset are provided in Table \ref{tab:cmv_modes_stats}. These stats are obtained after truncation of threads to 4096 token sequence length. During data analysis, we observed that several threads share the same initial post({\em submission}). Hence, we make sure that all threads with the same initial post entirely lie in either the train split, or the test. 

For both CMV Modes, and Dr. Inventor Corpus, we only consider contiguous spans of texts as single components, as opposed to the labelling in the dataset. Discontiguous spans are re-labelled as separate components and the model is trained and tested with these new labels, instead. 

For CMV Modes dataset, we add an extra "continue" class of relations to denote relation between two dis-contiguous spans of same argumentative component annotated in the data. We group together various relation types annotated in the CMV modes data into the 5 broad classes as follows: \textbf{support}("continue" class and "support" class), \textbf{agreement}("agreement", "understand" classes), \textbf{direct attack}("attack", "rebuttal attack", "rebuttal", "disagreement" classes), \textbf{undercutter attack}("undercutter", "undercutter attack" classes), \textbf{partial}("partial agreement", "partial attack", "partial disagreement" classes). These groupings are based on the broad annotation guidelines provided for the annotations of CMV Modes data.

For Dr. Inventor Corpus, due to the low number of \textbf{semantically same} relations(44) compared to \textbf{support}(4535) and \textbf{contradicts}(564) in the original dataset, we add the label("\textbf{parts-of-same}") which indicates that two dis-contiguous spans belong to the same argumentative component to the \textbf{semantically same} category. We also, merge together sections of papers to efficiently utilise 4096 token length of Longformer model. The detailed statistics after truncation to 4096 sequence length are presented in Table \ref{tab:dr_inv_stats}.
\begin{table}[]
    \footnotesize
    \centering
    \begin{tabular}{l|c}
        \hline
        {\bf Component Type} &  {\bf $\#$ Tokens}\\ \hline
         O          &   28186 \\
         B-C        &     1650  \\
         I-C        &    26529 \\
         B-P        &     1980 \\
         I-P        &    36552 \\
         \hline
         {\bf Relation Types}&  {\bf {$\#$ of relations}} \\ \hline
         support    &               1859 \\
         agreement  &                421 \\
         direct attack      &        283 \\
         undercutter attack &        330 \\
         partial            &        215 \\
        \hline
    \end{tabular}
    \caption{Statistics for the CMV-Modes dataset.}
    \label{tab:cmv_modes_stats}
\end{table}

\begin{table}[]
    \footnotesize
    \centering
    \begin{tabular}{l|c}
        \hline
        {\bf Component Type} &  {\bf $\#$ Tokens}\\ \hline
         O          &    153429 \\
         B-BC        &     3215  \\
         I-BC        &    39574 \\
         B-OC        &     5300 \\
         I-OC        &    74239 \\
         B-D        &     3994 \\
         I-D        &    19058 \\
         \hline
         {\bf Relation Types}&  {\bf {$\#$ of relations}} \\ \hline
         support    &               4535 \\
         Contradicts  &              564 \\
         Semantically Same      &   1049 \\
        \hline
    \end{tabular}
    \caption{Statistics for the Dr. Inventor dataset.}
    \label{tab:dr_inv_stats}
\end{table}
\section{Implementation Details}
\label{app:implement}

We use the pretrained base version of Longformer ($12$ layers, $768$ model size). The size of the local attention window was set to the default $512$. The maximum sequence length was fixed at $4096$. We added the special tokens that we used, to the pretrained Longformer tokenizer. For \textbf{ACI} our models use a CRF layer\footnote{We use the implementation of AllenNLP~\citep{gardner2018allennlp}}. \textbf{sMLM} training for Longformer based models was done on thread level and for BERT and RoBERTa based models on comment-level. We used mini-batch learning; approximately similar length input threads were batched together keeping the total number of tokens per batch fixed to $8,194$ for Longformer and $1024$ for BERT and RoBERTa models, and accumulated gradients over 3 batches. 

We trained our models for a total of $10$ epochs on sMLM task, while saving checkpoints after each epoch. We used Adam optimizer with a learning rate of $10^{-6}$. For all downstream tasks, we train our models for $30$ epochs, again, with Adam optimizer with learning rate of $2e-5$ as suggested by \citet{mosbach2021stability}. We use same batch sizes as \textbf{sMLM} training and accumulate gradients over 4 batches. We observe that for prompting {\bf RTP} on CMV-Modes, not making {\bf [USER-i]} tokens global, leads to better performance, hence we report results for same.

We find that \textbf{sMLM} training for 4 epochs is most beneficial, for performance on downstream task. Hence, we report results for the same checkpoint. Following the suggestions in \citet{reimers2017reporting}, we repeat our experiments on 5 different data splits and present the distributions in the Appendix. For the results at any epoch, the score plotted corresponds to mean over the 5 runs, and error regions correspond to the Bessel corrected standard deviation. The scores reported in the tables for various models correspond to the average value of the mean of 5 runs, over the last 5 epochs for that particular metric.
\begin{table}[!t]
    \scriptsize
    \centering
    \begin{tabular}{l|l}
        \multicolumn{1}{l|}{\bf Type} & \multicolumn{1}{c}{\bf Markers} \\
        \hline
        Opinion & {\em i agree, i disagree, i think, in my opinion, imo, imho}\\
        \hline
        Causation & {\em because, since, as, therefore, if, so,}\\ & {\em according to, hence, thus, consequently}\\
        \hline
        Rebuttal & {\em in contrast, yet, though, in spite of, but}\\& {\em regardless of, however, on the contrary}\\
        \hline
        Factual & {\em moreover, in addition, further to this,}\\& {\em in fact, also, firstly, secondly, lastly}\\
        \hline
        Assumption & {\em in the event of, as long as, so long as,}\\& {\em provided that, assuming that, given that}\\
        \hline
        Summary & {\em tldr}\\
        \hline
        Misc. & {\em why, where, what, how, when, while}\\
        \hline
    \end{tabular}%\vspace{-2mm}
    \caption{\footnotesize Types and examples of different discourse markers used for selective MLM finetuning.}\label{tab:markers}
\end{table}
Table~\ref{tab:markers} provides examples of markers of various kinds, that are masked during the {\bf sMLM} training.

\section{Additional results}
% We do not deal with the problem of predicting whether a relation exists between two component or not, as it is already done in \cite{ampersand}. We only try to predict the type of relations between components which are already labelled as related to one another in the dataset.

% \subsubsection{Main Results}
\begin{figure}[h]
    \centering
    \includegraphics[width=\columnwidth]{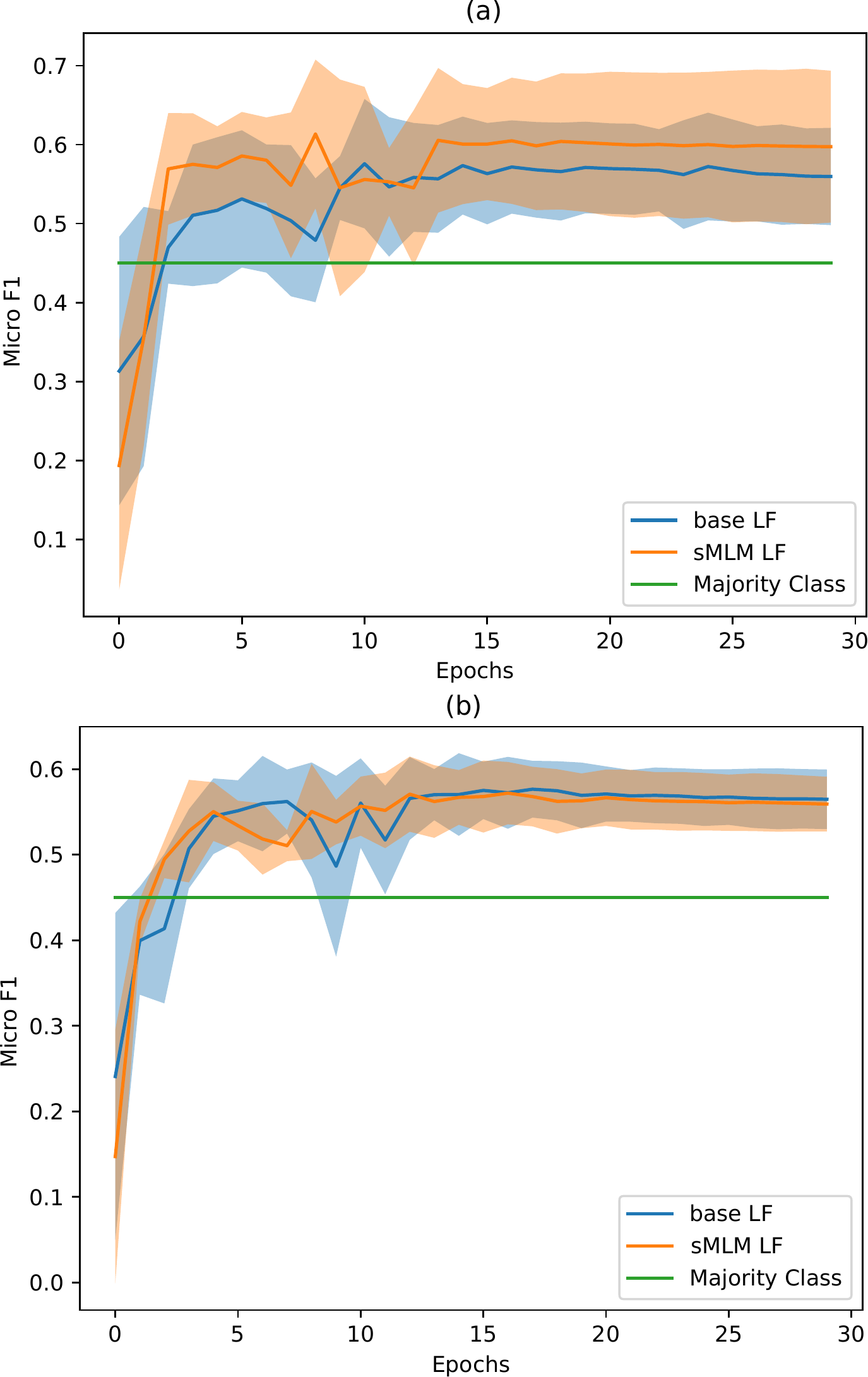}
    \caption{On CMV Modes data, sMLM-LF-mp's mean F1 converges to 0.59 compared to 0.56 for Base-LF-mp in 80-20 split (a) and 0.56 in 50-50 split (b).}
    \label{fig:mean_pooling_rel_pred_80_20}
\end{figure}

\begin{figure}
    \centering
    \includegraphics[width=\columnwidth]{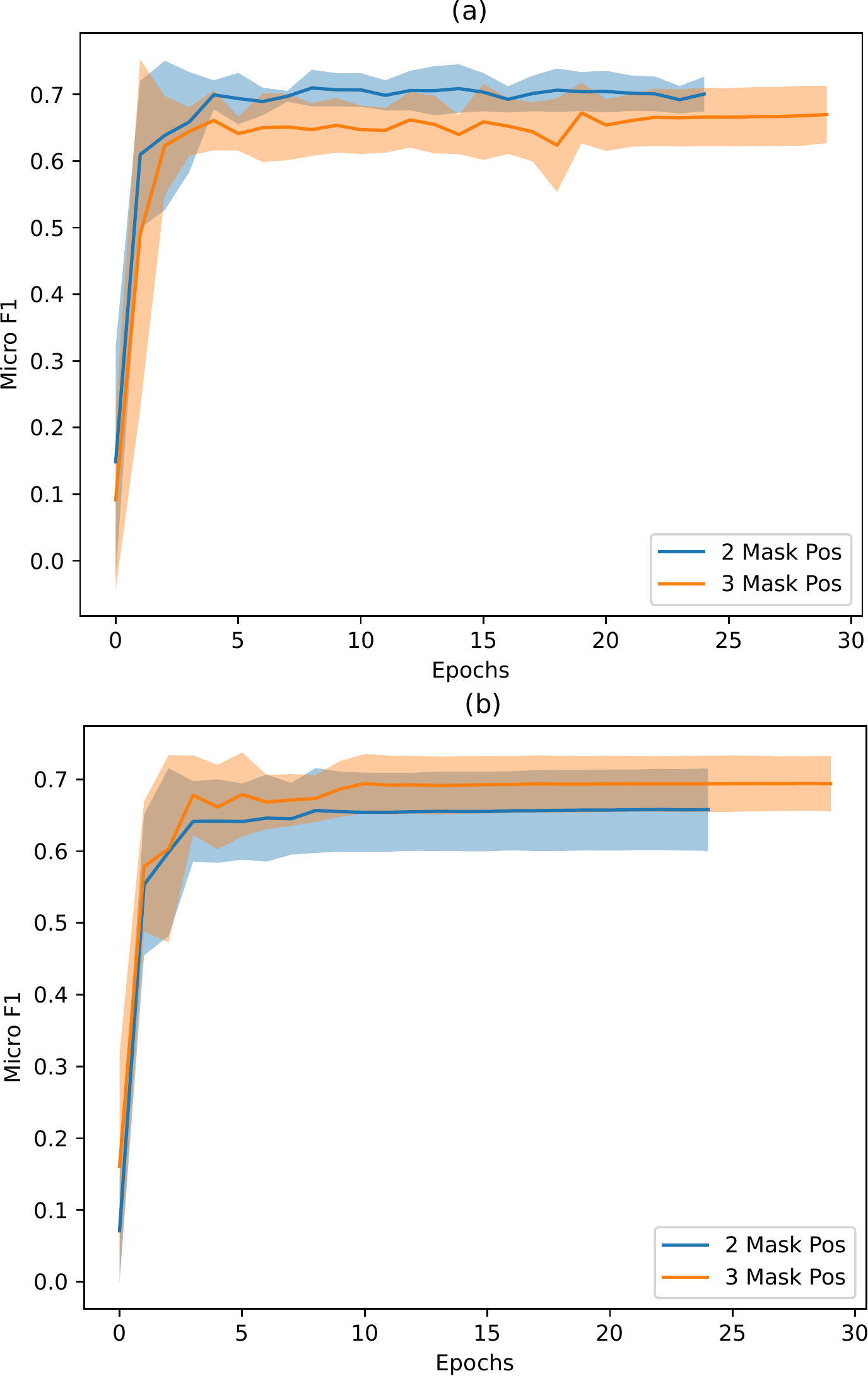}
    \caption{Change in sMLM-LF performance on CMV Modes {\bf RTP} (a) 80-20 and (b) 50-50 split when number of mask tokens in the prompt is changed from 3 to 2. The model with 2 masked token converges to 0.70 (0.66) and the mean for 3 masked tokens converges to 0.67 (0.69).}
    \label{fig:2_vs_3_mask_pos_smlm_lf}
\end{figure}

% \begin{figure}
%     \centering
%     \includegraphics[width=\columnwidth]{Figures/2_vs_3_mask_pos_ckpt4_1.pdf}
%     \caption{Change in sMLM-LF performance(on 50-50 split) when number of mask tokens changed from 3 to 2. The model with 2 masked token converges to 0.66 and the mean for 3 masked tokens converges to 0.69}
%     \label{fig:2_vs_3_mask_pos_smlm_lf1}
% \end{figure}

% \subsubsection{The Role of Context}
\begin{figure}
    \centering
    \includegraphics[width=\columnwidth]{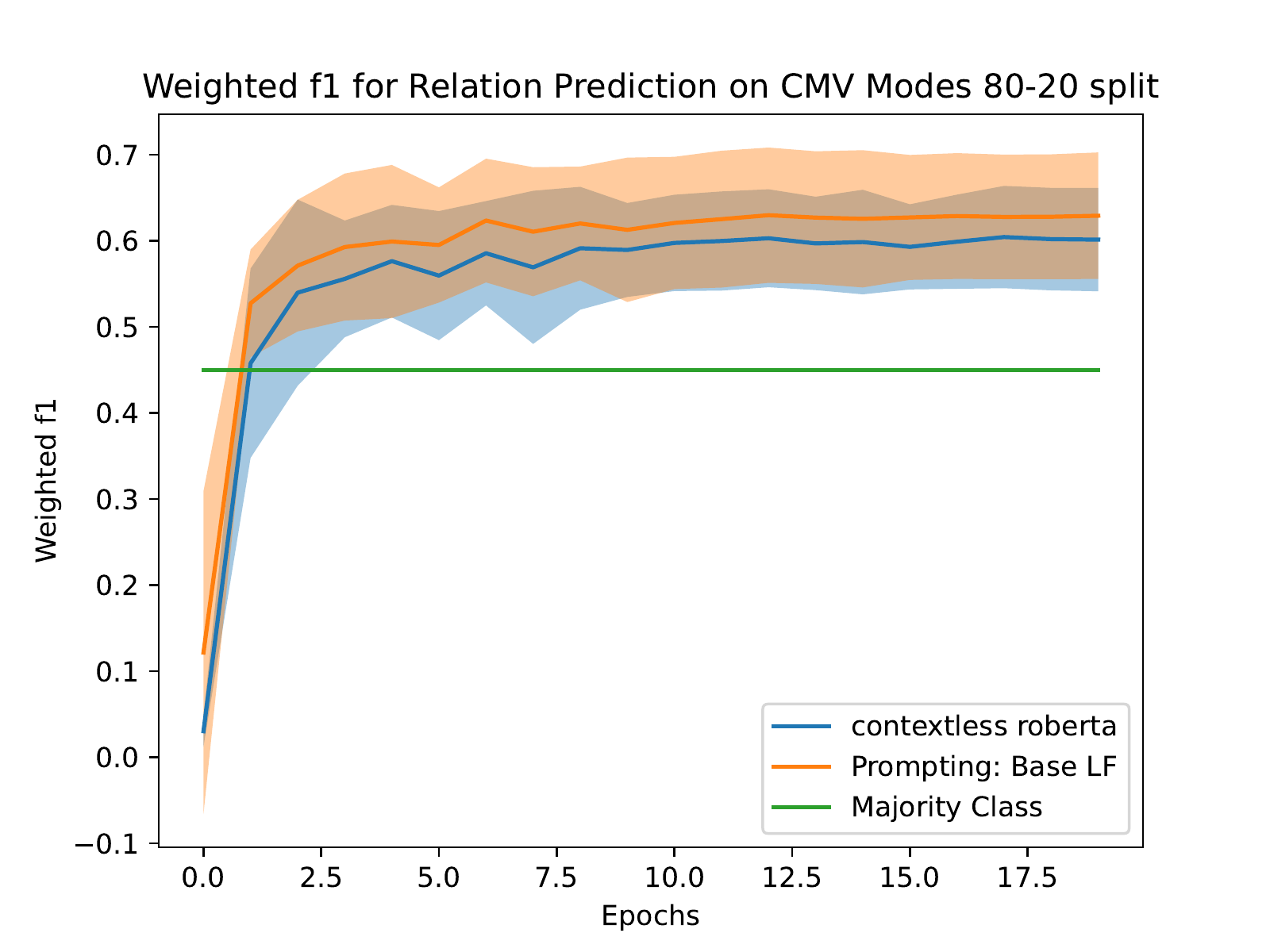}
    \caption{Contextless Roberta's mean f1 converges to around 0.599, compared to 0.62 of Base Longformer on {\bf RTP}.}
    \label{fig:contextless_roberta}
\end{figure}

\begin{figure}
    \centering
    \includegraphics[width=\columnwidth]{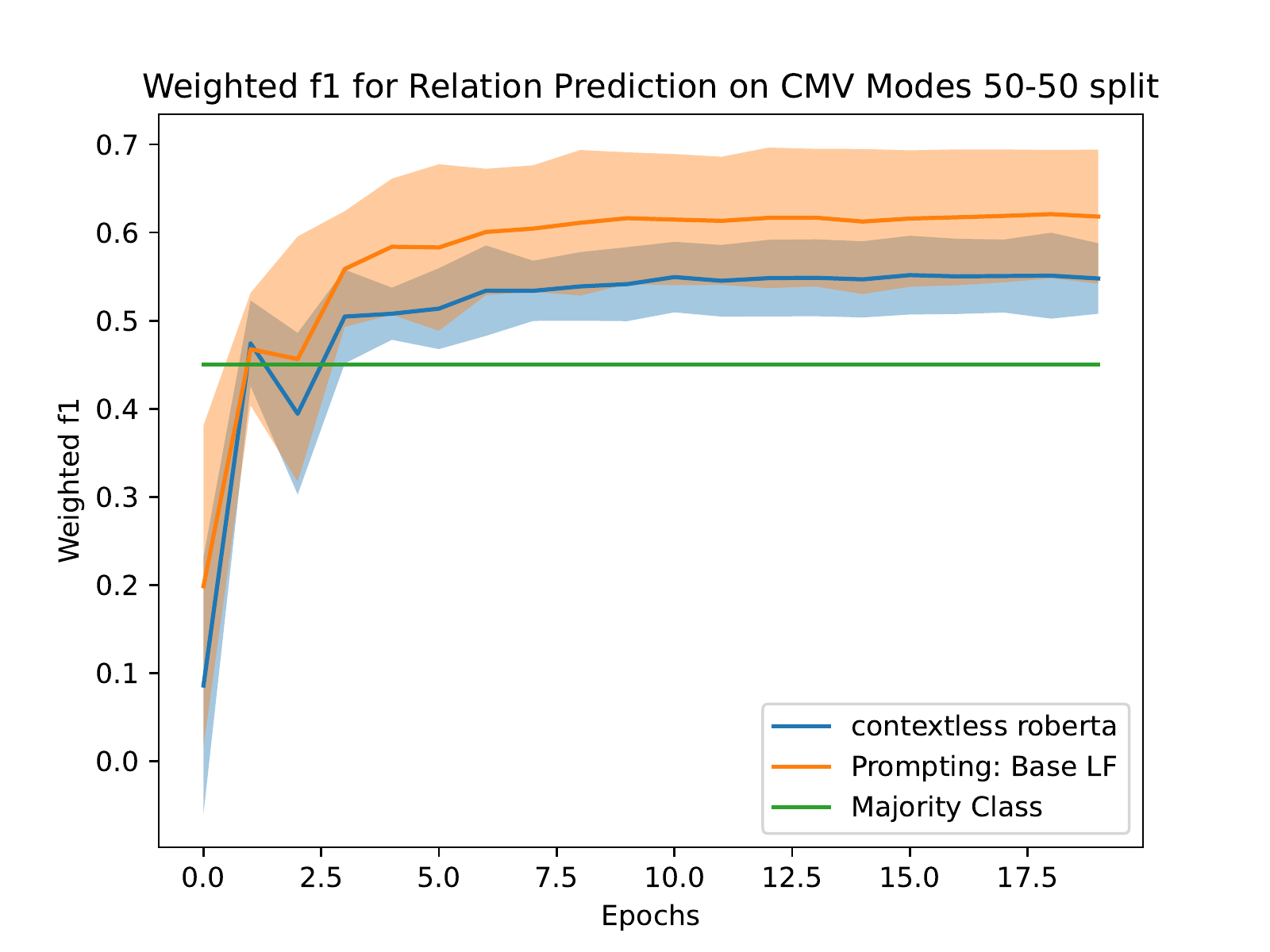}
    \caption{Contextless Roberta's mean f1 converges to around 0.55, compared to 0.617 of Base Longformer on {\bf RTP}.}
    \label{fig:contextless_roberta1}
\end{figure}
% Figure~\ref{fig:contextless_roberta} and \ref{fig:contextless_roberta1} show a comparison between the \textbf{contextless-RoBERTa} model and Base-LF trained with prompting strategy. As longformer is derived from RoBERTa itself, this allows us to investigate role of providing extra context. We observe an increase in performance on both splits. A comparison between {\bf QR-BERT} and {\bf contextless-RoBERTa} is provided in Figures~\ref{fig:contextless_qr_bert_80_20} and \ref{fig:contextless_qr_bert_50_50}.

\begin{figure}
    \centering
    \includegraphics[width=\columnwidth]{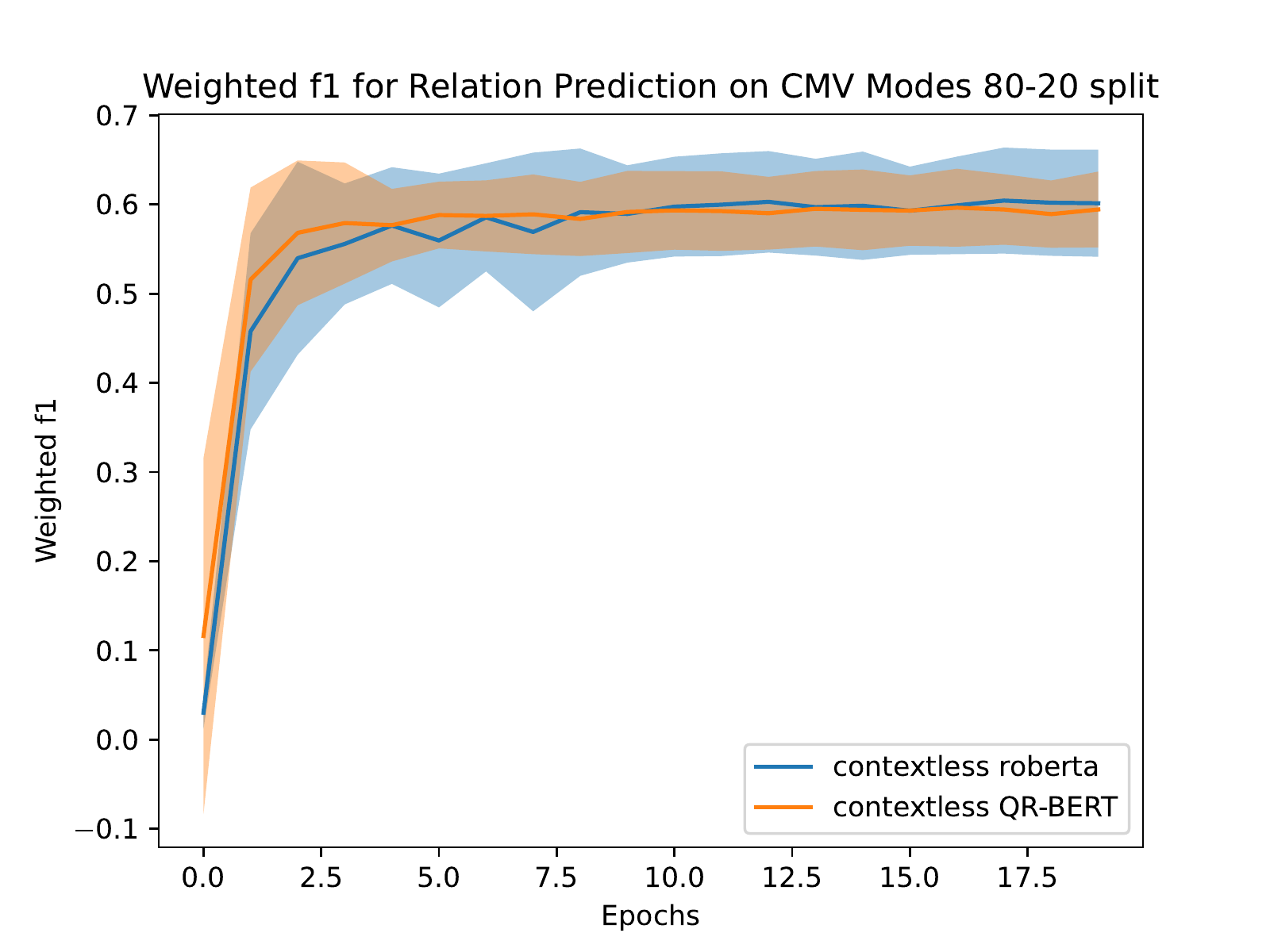}
    \caption{QR-BERT converges to an f1 score 0.59 compared to 0.60 for RoBERTa on the 80-20 split of CMV-Modes for {\bf RTP}.}
    \label{fig:contextless_qr_bert_80_20}
\end{figure}

\begin{figure}
    \centering
    \includegraphics[width=\columnwidth]{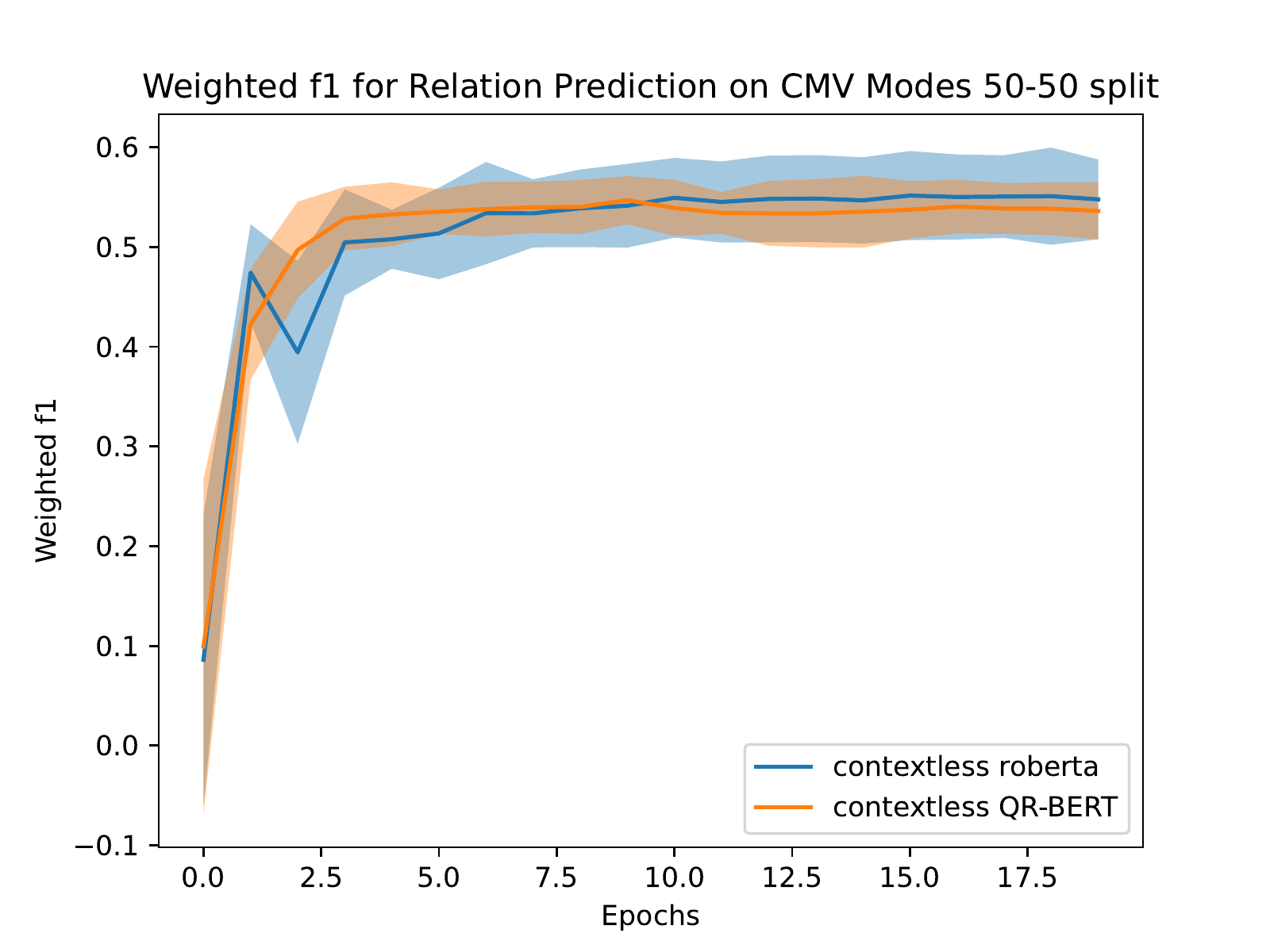}
    \caption{QR-BERT converges to an f1 score 0.54 compared to 0.55 for RoBERTa on the 50-50 split of CMV-Modes for {\bf RTP}.}
    \label{fig:contextless_qr_bert_50_50}
\end{figure}

% \subsubsection{Generalization to Other Datasets}

\begin{figure}
    \centering
    \includegraphics[width=\columnwidth]{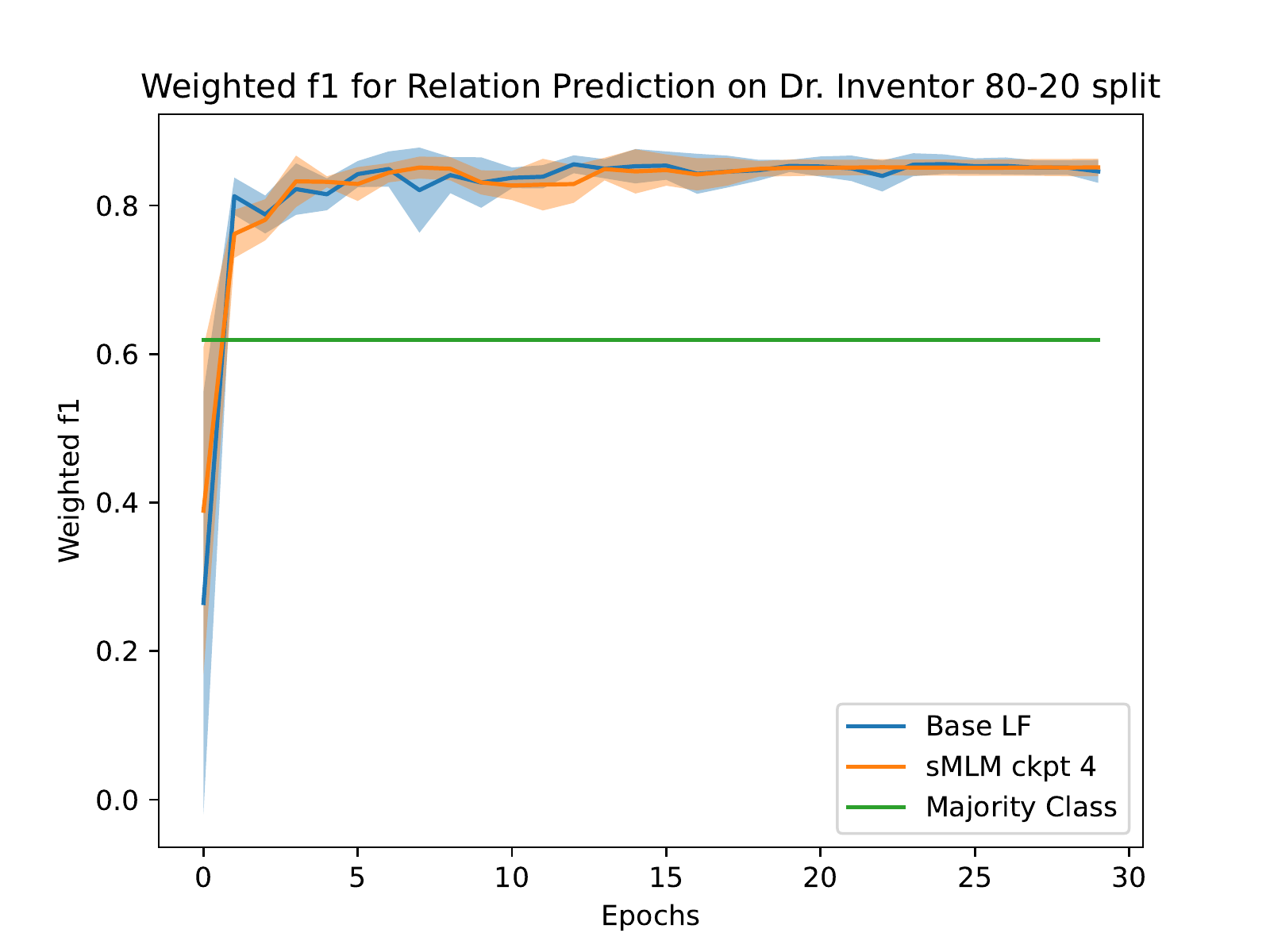}
    \caption{Both Base LF and our sMLM pretrained Longformer converge to an f1 of 0.85 with prompt-based {\bf RTP} on Dr. Inventor corpus.}
    \label{fig:prompt_dr_inv_80_20}
\end{figure}

% \begin{figure}
%     \centering
%     \includegraphics[width=\columnwidth]{Figures/prompt_dr_inv1.pdf}
%     \caption{Both Base LF(0.853) and our sMLM pretrained Longformer(0.844) converge to approximately same value}
%     \label{fig:prompt_dr_inv_50_50}
% \end{figure}

% To test whether our results hold under domain shift, we did prompt-based fine-tuning on the argument annotated\cite{ArgDrInvCorpus} Dr. Inventor Corpus\cite{DrInvCorpus}. We observe that the improvements in f1 scores don't carry over to this dataset. This is shown in Figures \ref{fig:prompt_dr_inv_80_20} \& \ref{fig:prompt_dr_inv_50_50}. It is expected, as the Winning Args Corpus\citep{MLMfinetune-data} is specific to \textbf{r/changemyview} subreddit. But still, we find that even under the domain shift the sMLM model manages to perform as good as Base Longformer, across different split sizes. 

% \subsection{Random Masking versus sMLM}
\begin{figure}
    \centering
    \includegraphics[width=\columnwidth]{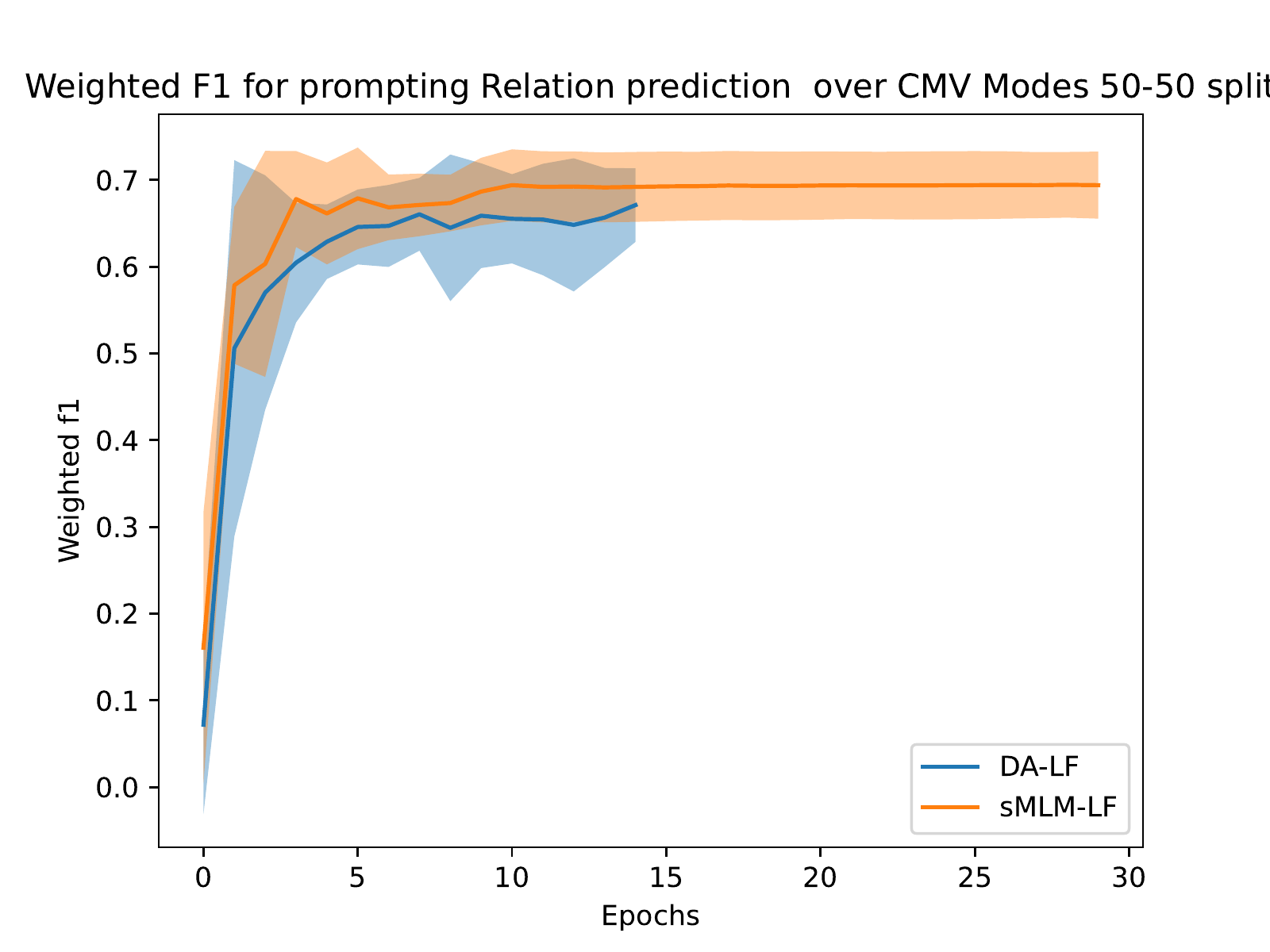}
    \caption{The Domain Adapted LF converges to around 0.66 compared to 0.69 for sMLM-LF, on the 50-50 split on CMV-Modes}
    \label{fig:da_vs_smlm1}
\end{figure}
\begin{figure}
    \centering
    \includegraphics[width=\columnwidth]{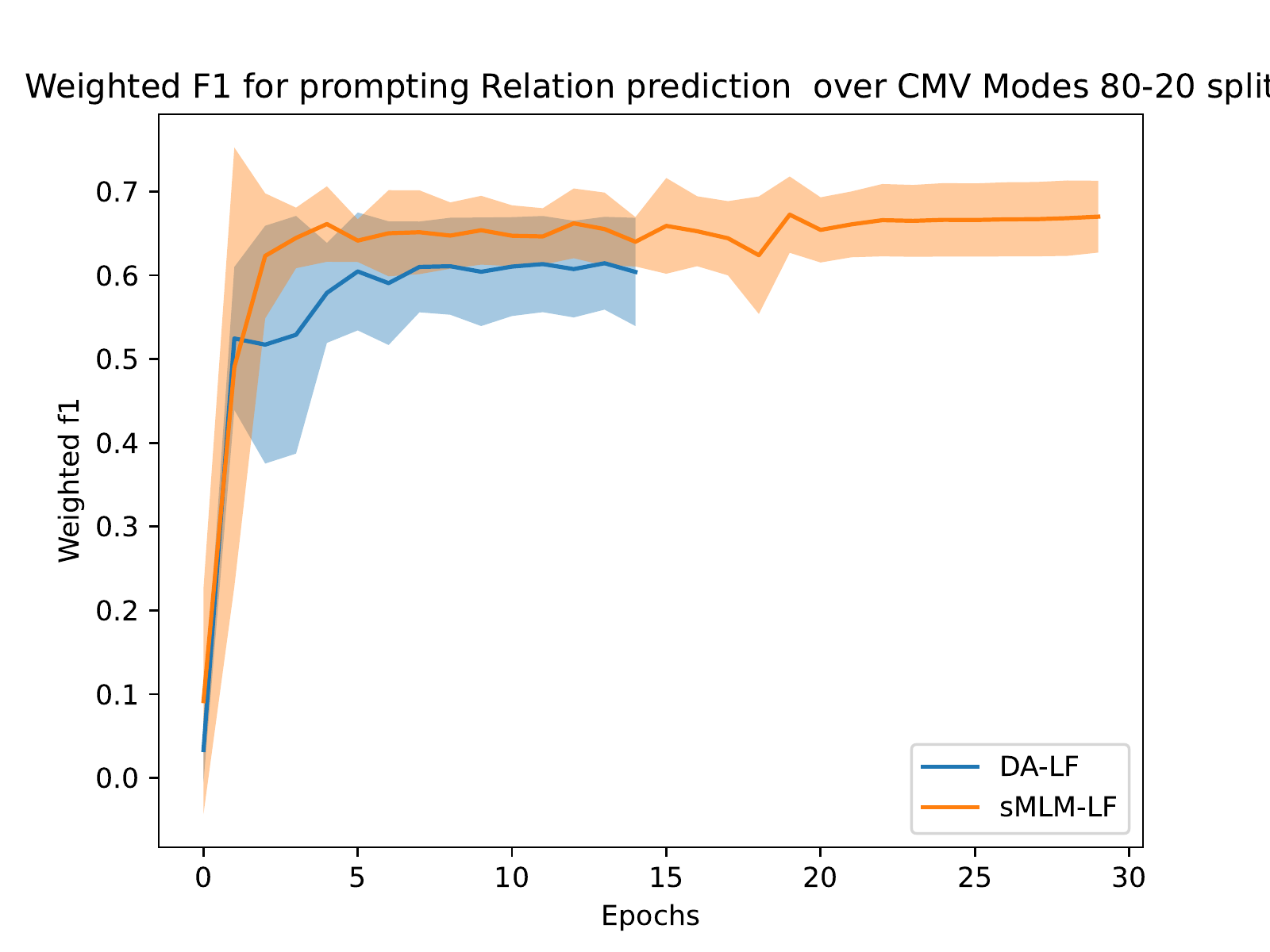}
    \caption{The Domain Adapted LF converges to around 0.61 compared to 0.67 for sMLM-LF, on the 80-20 split on CMV-Modes}
    \label{fig:da_vs_smlm}
\end{figure}
\end{document}